%% file: main.tex
\documentclass[twoside]{article}

\usepackage[accepted]{aistats2020}
\usepackage[round]{natbib}

\usepackage{tikz_prm}
\usepackage{pifont}
\newcommand{\cmark}{\ding{51}}%
\newcommand{\xmark}{\ding{55}}%

\usepackage[resetlabels,labeled]{multibib}

\newcites{App}{References}

\usepackage[utf8]{inputenc} 
\usepackage[T1]{fontenc}    
\usepackage[hidelinks]{hyperref}
\usepackage{url}            
\usepackage{booktabs}       
\usepackage{amsfonts}       
\usepackage{nicefrac}       
\usepackage{microtype}      
\usepackage[toc,page]{appendix}
\usepackage{algorithm}
\usepackage[noend]{algpseudocode}
\usepackage{stmaryrd}
\usepackage{amsmath, bm}
\usepackage{dsfont}
\usepackage{amssymb}
\usepackage{cancel}
\usepackage{ulem}
\usepackage{varwidth}
\usepackage{caption}
\usepackage{stackrel}
\usepackage{wrapfig}
\usepackage{xspace}
\usepackage{comment}

\usepackage{pgfplots}

\newcommand{\leftrarrows}{\mathrel{\raise.75ex\hbox{\oalign{%
  $\scriptstyle\leftarrow$\cr
  \vrule width0pt height.5ex$\hfil\scriptstyle\relbar$\cr}}}}
\newcommand{\lrightarrows}{\mathrel{\raise.75ex\hbox{\oalign{%
  $\scriptstyle\relbar$\hfil\cr
  $\scriptstyle\vrule width0pt height.5ex\smash\rightarrow$\cr}}}}
\newcommand{\Rrelbar}{\mathrel{\raise.75ex\hbox{\oalign{%
  $\scriptstyle\relbar$\cr
  \vrule width0pt height.5ex$\scriptstyle\relbar$}}}}

\makeatletter
\def\leftrightarrowsfill@{\arrowfill@\leftrarrows\Rrelbar\lrightarrows}
\newcommand{\xleftrightarrows}[2][]{\ext@arrow 3399\leftrightarrowsfill@{#1}{#2}}
\makeatother

\usepackage{xcolor}

\makeatletter
\def\mathcolor#1#{\@mathcolor{#1}}
\def\@mathcolor#1#2#3{%
  \protect\leavevmode
  \begingroup
    \color#1{#2}#3%
  \endgroup
}
\makeatother
\colorlet{darkgreen}{green!60!blue}

\newcommand{\nn}{n}
\newcommand{\oo}{o}
\newcommand{\pp}{p}
\newcommand{\oap}{op}

\newcommand{\om}{{n_-}}
\newcommand{\op}{{n_+}}
\newcommand{\mint}{\textstyle\int}

\newcommand{\ssvgp}{S$^2$VGP\xspace}
\newcommand{\svgp}{SVGP\xspace}

\newcommand{\vX}{ \boldsymbol{X} }
\newcommand{\vA}{ \boldsymbol{A} }
\newcommand{\vI}{ \boldsymbol{I} }
\newcommand{\vx}{ \boldsymbol{x} }
\newcommand{\vu}{ \boldsymbol{u} }

\newcommand{\vv}{ \boldsymbol{v} }
\newcommand{\vy}{ \boldsymbol{y} }
\newcommand{\vz}{ \boldsymbol{z} }
\newcommand{\vk}{ \boldsymbol{k} }
\newcommand{\vK}{ \boldsymbol{K} }

\newcommand{\vW}{ \boldsymbol{W} }
\newcommand{\vm}{ \boldsymbol{m} }
\newcommand{\vq}{ \boldsymbol{q} }
\newcommand{\vxi}{ \boldsymbol{\xi} }
\newcommand{\vSigma}{ \boldsymbol{\Sigma} }
\newcommand{\vmu}{ \boldsymbol{\mu} }
\newcommand{\vs}{ \boldsymbol{s} }
\newcommand{\vP}{ \boldsymbol{P} }
\newcommand{\vT}{ \boldsymbol{T} }
\newcommand{\vt}{ \boldsymbol{t} }
\newcommand{\vR}{ \boldsymbol{R} }
\newcommand{\veps}{ \bm{\varepsilon} }
\newcommand{\vPhi}{ \boldsymbol{\Phi} }

\newcommand{\vQ}{ \boldsymbol{Q} }
\newcommand{\vM}{ \boldsymbol{M} }

\newcommand{\GP}{ {\cal GP} }
\newcommand{\vH}{ \boldsymbol{H} }
\newcommand{\vF}{ \boldsymbol{F} }
\newcommand{\vL}{ \boldsymbol{L} }
\newcommand{\cD}{ {\cal D} }
\newcommand{\cL}{ {\cal L} }
\newcommand{\vU}{ \boldsymbol{U} }
\newcommand{\order}{\mathcal{O}}
\newcommand{\XX}{ {\cal X} }
\newcommand{\NN}{ {\cal N} }
\newcommand{\RR}{ \mathds{R} }
\newcommand{\msum}{\textstyle \sum}
\newcommand{\EE}{ \mathds{E} }
\newcommand{\VV}{ \mathds{V} }

\newcommand*{\KL}[2]{\ensuremath{\operatorname{KL}[#1 \,\|\, #2]}}

\newcommand{\trace}{\operatorname{tr}}
\newcommand{\cov}{\operatorname{cov}}
\newcommand{\dee}{\mathrm{d}}

\newcommand{\btd}{\operatorname{btd}}
\newcommand{\chol}{\operatorname{chol}}
\newcommand{\mprod}{\textstyle\prod}
\newcommand{\bigO}{\mathcal{O}}

\newcommand{\vince}[1]{\textcolor{black}{#1}}  
\newcommand{\stef}[1]{\textcolor{black}{#1}} 

\newcommand*\rot{\rotatebox{0}}

\begin{document}

\twocolumn[

\aistatstitle{Doubly Sparse Variational Gaussian Processes}

\aistatsauthor{ Vincent Adam \And Stefanos Eleftheriadis \And Nicolas Durrande \And Artem Artemev \And James Hensman }

\aistatsaddress{ PROWLER.io, Cambridge, UK} ]

\begin{abstract}
The use of Gaussian process models is typically limited to datasets with a few tens of thousands of observations due to their complexity and memory footprint.
The two most commonly used methods to overcome this limitation are 1) the variational sparse approximation which relies on inducing points and 2) the state-space equivalent formulation of Gaussian processes which can be seen as exploiting some sparsity in the precision matrix.
We propose to take the best of both worlds: we show that the inducing point framework is still valid for state space models and that it can bring further computational and memory savings. Furthermore, we provide the natural gradient formulation for the proposed variational parameterisation.
Finally, this work makes it possible to use the state-space formulation inside deep Gaussian process models as illustrated in one of the experiments. 

\end{abstract}

\setlength{\abovedisplayskip}{5pt}
\setlength{\belowdisplayskip}{5pt}
\setlength{\abovedisplayshortskip}{0pt}
\setlength{\belowdisplayshortskip}{0pt}

\section{Introduction}
Gaussian processes (GPs) provide a very powerful framework for statistical modelling in low data regimes (i.e., when the number of observations $N$ is small).
However, they typically scale as $\order{(N^3)}$ in computational complexity and $\order{(N^2)}$ in memory which makes them impractical for datasets containing more than a few thousand observations. 
This limitation has received a lot of attention, especially in the machine learning community~\citep{Rasmussen2006}, and two frameworks have distinguished themselves so far. 

The first approach focuses on state-space models (SSM) and exploits the underlying Markov property for computational efficiency. The Markov structure results in sparse precision matrices \citep{grigorievskiy2017parallelizable, durrande2019banded}, or enables filtering algorithms \citep{kalman1960, sarkka_solin_2019, solin2018infinite}. These approaches lead to inference algorithms with $\order{(N)}$ complexity. 
Several model types can be tackled with this approach, such as Gaussian Markov random fields or linear stochastic differential equations. 
The second approach is the sparse variational Gaussian process (\svgp). 
It relies on the assumption that the {dataset $\cD =\{x_i, y_i\}_{i=1}^N$ contains some redundant information, so that the GP posterior $f | \{f(x_i)=y_i\}_{i=1}^N$ can be approximated using a smaller set of inducing points (i.e.\ pseudo-inputs): $f | \{f(z_i)=u_i\}_{i=1}^M$ with $M \ll N$}.
Variational Inference (VI), which consists in minimising the Kullback--Leibler divergence between the approximate and the true posterior, is then used to find appropriate values for $\vz$, for and for the model parameters \citep{titsias2009variational, hensman2013gaussian}.

\begin{figure}[t!]
    \includegraphics[width=.49\textwidth]{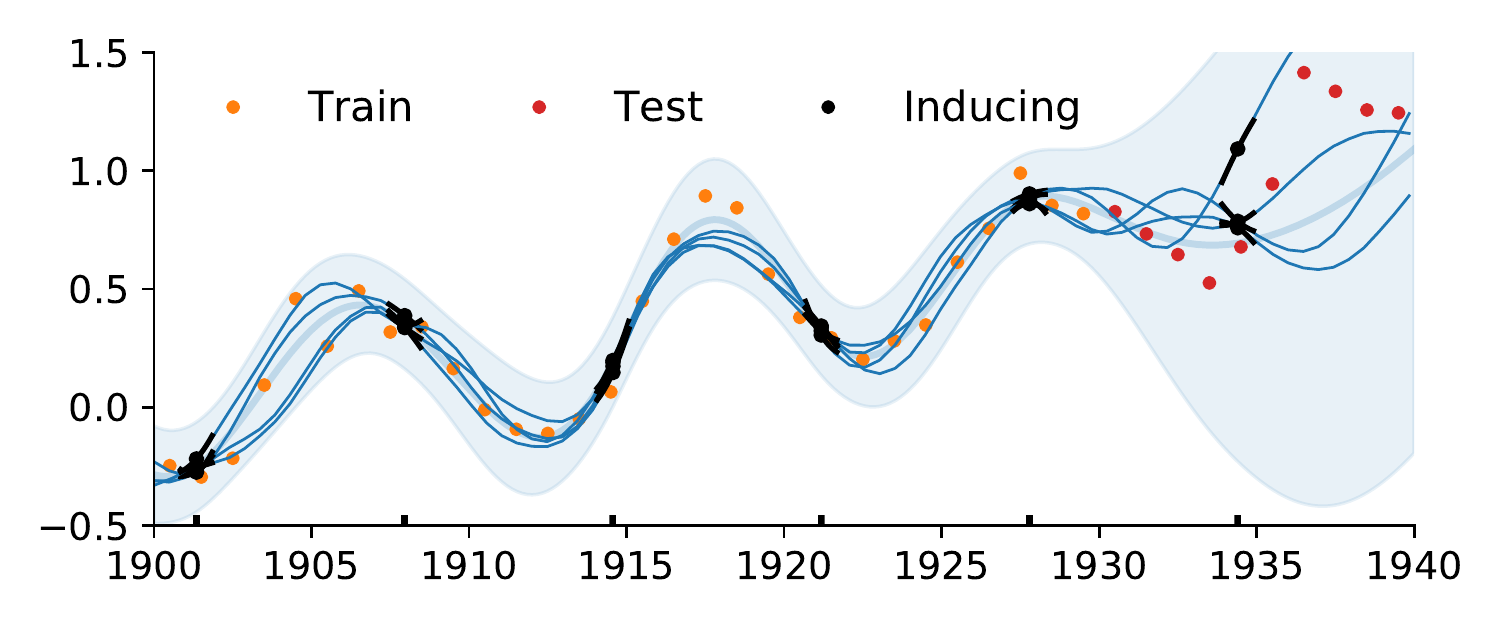}
    \caption{Illustration of the proposed method for a regression task with a Mat\'ern \nicefrac{5}{2} kernel. For each inducing location $z_i$, we introduce three inducing variables, i.e., $f(z_i),\ f'(z_i)$ and $f''(z_i)$. Figure also shows samples from the model, where the inducing variables are represented by a quadratic Taylor expansion. For more details on the experiment settings see Section 4.1.} \label{fig:first_example}
\end{figure}

In this paper, we focus on GPs with 1-dimensional input and propose a novel inference method that brings together the merits of both sparse GP approximation and SSM representation: for a state-space model with state dimension $d$, we introduce $d$-dimensional inducing variables and associate each element of this vector to a state component. For example, a GP $f$ with a Mat\'ern \nicefrac{5}{2} covariance is Markovian if you consider the state space $(f(x), f'(x), f''(x))$. Given some inducing locations $z_i$, we approximate the posterior $f | \{f(x_i)=y_i\}_{i=1}^N$, by $f | \{f(z_j)=u_{j, 1}, f'(z_j)=u_{j, 2}, f''(z_j)=u_{j, 2}\}_{j=1}^M$ (see Figure~\ref{fig:first_example}).
Using the state space components as inducing features brings several advantages. First, the proposed method scales linearly both with the number of data and with the number of inducing points (whereas \svgp is quadratic in the later). Second, the number of variational parameters that need optimising is $\order{(Md^2)}$, which scales favourably compared to $\order{(M^2)}$ and $\order{(Nd^2)}$ as respectively required by \svgp and VI for SSM \citep{durrande2019banded}. Finally, our approach allows for mini-batching as well as the use of SSM layers in deep-GP models.

In developing our approach, we quickly identified that optimisation of the objective function was 
cumbersome using standard gradient approaches. We developed a natural gradient approach for our method based on \cite{salimbeni18a}. To do this efficiently requires the formulation of the compact exponential family form of a Markov structured Gaussian distribution, with novel mathematical and computational operators.

\section{Background}
Here we introduce the basics of GP models~\citep{Rasmussen2006} as well as the two main techniques for dealing with large datasets: sparse variational inference and inference in the state-space formulation.
They both lead to sparse algorithms, albeit in a different way.

\subsection{Gaussian processes}

In the classic GP regression setting we are given a dataset $\cD =\{\vx_n, y_n\}_{n=1\dots N} \in (\XX , \RR)^N$, 
where $y_n$ corresponds to the evaluation of a latent function corrupted with observation noise $y_n = f(\vx_n) + \epsilon_n$
and the task is to predict $f(\vx_{*})$ for some input location $\vx_{*} \in \XX$.
Under the assumptions that $f$ is a Gaussian process $f(\cdot)\sim \GP(0, k(\cdot, \cdot))$ and that $\epsilon$ also follows a multivariate normal distribution $\epsilon \sim \NN(0, \sigma^2 \vI)$, the prediction problem can be solved analytically and results in a Gaussian distribution:
\begin{align}
    \EE_{f|\cD}[f(\vx_*)] &= \vk_{*}^{\top}(\vK + \sigma^2 \vI)^{-1}\vy \\
    \VV_{f|\cD}[f(\vx_*)] & = k(\vx_{*},\vx_{*}) - \vk_{*}^{\top}(\vK + \sigma^2 \vI)^{-1}\vk_{*}\,,
\end{align}
where the matrix $\vK$ is defined as $\vK_{ij}=k(\vx_i,\vx_j)$ and $\vk_{*}$ is vector such that $\vk_{*i} = k(\vx_{*}, \vx_i)$.
The matrix inverse required in the prediction yields a computational complexity of $\order(N^3)$ which is prohibitively expensive for large datasets.

\subsection{Variational inter-domain approximations}

Variational inference in probabilistic models turns inference into an optimisation problem \citep{jordan1999introduction}.
The sparse variational Gaussian process, originally introduced by \citet{titsias2009variational} and more rigorously defined by \citet{matthews2016sparse},
approximates the posterior distribution $p(f(\cdot)|\cD)$ by a distribution $q(f)$ that depends on `inducing points':
\begin{align}
q(f(\cdot))= \mint p(f(\cdot) | \{f(\vz_i) = \vu_i\}_{i=1}^M)q_{\vu}(\vu) \dee \vu\,.
\label{eq:sparseapprox}
\end{align} 
The inducing variables $\vu$ are typically assumed to be normally
distributed (i.e.,\ $q_{\vu} =\NN(\vmu_{\vu}, \vSigma_{\vu\vu})$) with the moments corresponding to the variational parameters.

The above approach has been generalised by introducing the idea of inter-domain
features \citep{alvarez2009sparse, lazaro2009inter} which replaces the conditioning $\{f(\vz_i) = \vu_i\}$ in Eq.~\eqref{eq:sparseapprox} by $\{\Psi_i[f(\cdot)] = \vu_i\}$ with $\Psi$ a linear operator $\RR^\XX \to \RR$ .
Choosing $\Psi_i: f \to f(\vz_i)$ allows to recover the classic inducing points but more interesting behaviour can be obtained by using other operators such as the integrals or convolutions \citep{hensman2017variational, vdwilk2017convolutional}.
To enhance readability, we denote by $\Psi[f]$ the vector of size $M$ with entries $\Psi_i[f]$, and by $p_\Psi$ its distribution.

The distribution of the conditioned GP is
\begin{align}
\label{eq:inter-dom-cond}
    f(\cdot)| \{ &\Psi[f] = \vu \} \sim \\
    &\nonumber\NN \left(\vk_{\Psi}^\top(\cdot) \vK_{\Psi \Psi}^{-1}\vu,\, k(\cdot, \cdot) - \vk_{\Psi}^\top(\cdot)\vK_{\Psi \Psi}^{-1}\vk_{\Psi}(\cdot)\right)\,,
\end{align} 
with $(\vk_\Psi(x))_i = \cov(f(x), \Psi_i[f])$ and $(\vK_{\Psi\Psi})_{i,j} = \cov(\Psi_i[f], \Psi_j[f])$. Taking the expectation of Eq.~\eqref{eq:inter-dom-cond} under $q_{\vu}(\vu)$ gives a closed form expression for $q(\cdot)$:
\begin{align}
    q(\cdot) &= \GP \big(\vk_{\Psi}^\top(\cdot) \vK_{\Psi \Psi}^{-1}\vmu_{\vu},\,\\
    \nonumber& k_{ff}(\cdot, \cdot) - \vk_{\Psi}^\top(\cdot)\vK_{\Psi \Psi}^{-1}(\vK_{\Psi \Psi} - \vSigma_{\vu \vu})\vK_{\Psi \Psi}^{-1}\vk_{\Psi}(\cdot)\big).
\end{align}

The variational lower bound to the log-marginal likelihood is then given by:
\begin{align}
\label{eq:elbo}
\cL(q) = \EE_q [\log p(\vy|f(\vX))] - \KL{q_{\vu}}{p_{\Psi}}\,.
\end{align}
The evaluation of $\cL$ can be shown to scale as $\order(NM^2 + M^3)$.
The linear scaling with the size of the training set allows to apply this approximation to large datasets \citep{hensman2013gaussian}.
In some cases, it is shown to approximate the posterior process with high accuracy at a low computational cost in the large data regime \citep{burt2019}.
Finally, it provides a well-defined objective, amenable to gradient-based optimisation that works well in practice \citep{bauer2016}.

\subsection{Inference in the state-space formulation}
A large class of Gaussian processes defined on $\XX\subseteq\RR$ can be written as linear stochastic differential equations (SDEs). 
This class of kernels is described in depth in Chapter 12 of \citet{sarkka_solin_2019} and includes, for example, all Mat\'ern \nicefrac{k}{2} (k odd), harmonic oscillators, and all sum and products of such kernels. Also, many kernels can be approximated by an element of this class, see for example \cite{sarkka2014convergence} for the RBF kernel.

Here we focus on the class of GPs with stationary kernels which can be represented as a linear time-invariant (LTI) SDE of the form:
\begin{align}
    \dot{\vs}(t) &= \vF \vs(t) + \vL \veps(t)\,, \qquad
    f(t) = \vH \vs(t)\,.
\end{align}
The state-space vector $\vs(t)\in \RR^d$ is given by evaluations of the process and its derivatives $\vs(t) = [f(t), f^{(1)}(t), \dots, f^{(d-1)}(t)]^\top$.
$\veps(t) \in \RR^r$ is a white noise process
with spectral density $\vQ_c$. $\vF \in \RR^{d\times d}$, $\vL\in \RR^{d\times r}$, $\vH\in \RR^{1\times d}$ are the feedback,
noise effect and observation matrix of the system.

The marginal distribution of the solution of this LTI-SDE evaluated at any ordered set
$\vx = [x_1,\dots, x_N]^\top\in \RR^N$ follows a discrete-time linear system:
\begin{align}
    \vs(x_{n+1}) &= \vA_{n,{n+1}} \vs(x_n) + \vq_n\,, \quad \vq_n \sim \NN(0, \vQ_{n,{n+1}}) \,\nonumber\\
    \vs(x_0) &\sim \NN(0, \vP_0)\,, \quad f(x_n) = \vH \vs(x_n)\,
\end{align}
where the state transition matrices $\vA_{n,{n+1}} \in \RR^{d\times d}$, noise covariance matrices $\vQ_{n,{n+1}} \in \RR^{d\times d}$, and state stationary covariance matrix $\vP_0$
can be computed analytically.
Denoting $\vPhi$ the matrix exponential and $\Delta_n = x_{n+1} - x_{n}$, we have
\begin{align}
    \vA_{n,{n+1}} &= \vPhi( \vF \Delta_n)\,, \qquad\\
    \vQ_{n,{n+1}} &=  \mint_{0}^{\Delta_n} \vPhi(\Delta_n-\tau)\vL \vQ_c \vL^{\top} \vPhi(\Delta_n-\tau)^{\top} \dee\tau\,.
\end{align}

By following the standard Kalman recursions \citep{sarkka2013bayesian}, inference in conjugate models using the
state-space formulation of the GP prior scales linearly with
the number of data points and cubically with the state-space dimension, i.e., $\order(Nd^3)$.
Equally efficient approximations have been derived in the non-conjugate case \citep{durrande2019banded, nickisch2018state},
by exploiting the block-tridiagonal structure of the precision \citep{grigorievskiy2017parallelizable}.

\section{Doubly sparse inference}

In this section we introduce the idea of combining the inter-domain inducing features with the state-space GP formulations,
which results in what we dub a ``doubly sparse variational GP'' approximation (\ssvgp).
Although the scope of our approach is broader, we formulate our idea under the classic GP regression setting
with factorising likelihood:  $p(\vy, f(\cdot)|\vx) = p(f(\cdot))\prod_n p(y_n|f(x_n))$.

\subsection{State-space inducing features}\label{sec:ssm_features}
We restrict our analysis to GPs with an LTI-SDE representation
and we choose
$\Psi_i: f \to \vs(z_i) = [f(z_i), \dots, f^{(d-1)}(z_i)]^\top$
as our inter-domain features evaluated at \emph{ordered} inducing inputs $\vz= [z_1,\dots,z_M]^\top$.
This has two immediate desirable consequences:
\paragraph{Property 1\normalfont :}
The sequence of inducing states $\vu = \{\Psi_i[f]\}_{i=1}^M$ is Markovian and its distribution is multivariate normal $p_\Psi = \mathcal{N}(0, \vQ_{\Psi}^{-1})$. 
This means that $\vQ_{\Psi}$ has a block-band structure and that the sufficient statistics are {$\vt(\vu) = [\vu, \btd[\vu \vu^\top]]$},
where $\btd$ extracts the block-tridiagonal elements (with block size $d$) and returns them as a vector. This property can be summarised as:
\begin{align}
    p_\Psi(\vu) &= p(\vu_1) \mprod_m p(\vu_{m+1} | \vu_{m}) \propto \exp\left( \boldsymbol\theta^\top \vt(\vu)  \right)\,,
\end{align}
where $\boldsymbol \theta = [\vQ_{\Psi} \vmu_\Psi,
\nicefrac{-1}{2} \btd[\vQ_{\Psi}]]$ are the natural parameters of the distribution.

\begin{figure}[ht]
\input{graphs/graphical_model_small.tex}
\caption{\textbf{(a)} Graphical model representing the Markovian joint prior of states indexed at $\{x_n,z_m\}$ (black arrows). 
\textbf{(b)} Graphical model for marginal posterior $q(\vs_2,\vu)=p(\vs_2|\vu)q(\vu)$ highlighting the statistical properties of the variational posterior: \textbf{(1)} $q_{\vu}(\vu)$ is a Markovian sharing the same structure as $p_\Psi(\vu)$ (red arrows); \textbf{(2)} $\vs_2$ only depends on the two nearest inducing states, i.e., $p(\vs_2|\vu)=p(\vs_2|\vu_1,\vu_2)$ (blue arrows). The conditional dependencies  $p(\vs_i|\vu)$, are shown in light blue.} 
\label{fig:graph_model}
\end{figure}
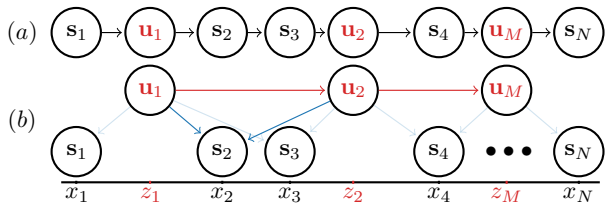

\paragraph{Property 2\normalfont :}
The posterior of the function evaluation at a point $x_n$ conditioned on the inducing variables,
depends only on the closest left and right inducing states:
\begin{align}
\nonumber    &p(f(x_n)|\vu) \\  
    \nonumber&= \frac{ \cancel{p(\vu_{1:\om})} 
    p(f(x_n)|\vu_{\om})p(\vu_{\op}|f(x_n))
    \cancel{p(\vu_{\op+1:M}|\vu_{\op})}}
    {\cancel{p(\vu_{1:\om})}p(\vu_{\op}|\vu_{\om}) \cancel{p(\vu_{\op+1:M}|\vu_{\op})}}\\
    &= p(f(x_n)|\vu_{\om},\vu_{\op})\, ,
\end{align}
where the indices $\om \in \llbracket 1,M-1 \rrbracket$ and $\op=\om+1$ correspond 
to the indices of the closest lower and upper neighbor of $x_n$ in $\vz$, i.e., $z_1<\dots < z_\om< x_n <z_\op <\dots < z_M$. 
This follows directly from $f(x_n) = \vH \vs (x_n)$ and the Markovian property of $\{ \vu_1, \dots, \vu_\om,  \vs(x_n) , \vu_\op, \dots, \vu_M\}$ which ensures that $p(\vs(x_n)|\vu) = p(\vs(x_n)|\vu_{\om}, \vu_{\op})$.
The graphical model associated with these properties is given in Figure~\ref{fig:graph_model}.
In practice, we get the statistics of $p(\vs(x_n)|\vu_{\om}, \vu_{\op})$ in time $\bigO(d^3)$ using the state-space parameters:
\begin{align}
\nonumber     p(\vs(x_n)|\vu_\om, \vu_\op) &\propto \NN(\vs(x_n); \vA_{\om,n} \vu_\om, \vQ_{\om,n})\\
&\times \NN(\vu_\op | \vA_{n,\op} \vs(x_n), \vQ_{n,\op}) \nonumber \\
     &= \NN(\vs(x_n); \vP_n \vv_n, \vT_n)\,,
\end{align}
where $\vv_n = [\vu_\om; \vu_\op]$. The matrices $\vP_n$, $\vT_n$ depend on the statistics of the prior state transitions between time points triplet ($z_\om, x_n, z_\op$) and are given in Appendix \ref{sec:appendix_conditional}.

Finally,  we define the marginal posterior on $\vv_n$ as $q(\vv_n) = \NN(\vmu_{\vv_n} , \vSigma_{\vv_n\vv_n})$, 
and obtain the posterior predictions analytically as
\begin{align}
\label{eq:marginal-prediction}
q(\vs(x_n)) =  \NN(\vs(x_n); \vP_n \vmu_{\vv_n}, \vT_n + \vP_n\vSigma_{\vv_n\vv_n}\vP_n^\top)\,.
\end{align}

\subsection{Optimal variational distribution for inducing state-space features}
\label{sec:variatonal_inf}

We continue our analysis by investigating the form of the approximating distribution on the inducing states 
$q_{\vu} = \NN(\vmu_{\vu}, \vSigma_{\vu\vu})$.
By following the approach of \citep{opper2009variational}, we show that at the optimum the variational distribution $q_{\vu}$
has a precision with the same block-tridiagonal structure as the prior precision $\vQ_\Psi$. More specifically, we start from the variational
loss in Eq.~\eqref{eq:elbo} and expand the terms of the KL that depend on the posterior covariance $\vSigma_{\vu\vu}$:
\begin{align}
\nonumber\cL(q) &= \msum_n \EE_{q} \left[\log p(y_n| f(x_n))\right] - \KL{q_{\vu}}{p_{\Psi}}\\
 &= \msum_n \EE_{q} \left[\log p(y_n| f(x_n))\right] - \frac{1}{2} \trace(\vQ_\Psi\vSigma_{\vu\vu}) \nonumber \\
 &+ \frac{1}{2}|\vSigma_{\vu\vu}| +c(\vmu_{\vu}, p_\Psi)\,,
\end{align}
where $c(\vmu_{\vu}, p_\Psi)$ contains the terms in KL that do not depend on $\vSigma_{\vu\vu}$.
At the optimal covariance $\vSigma^*_{\vu\vu}$, the gradient of the loss w.r.t. the variational parameters is zero, i.e. 
$\nabla_{\vSigma_{\vu\vu}}\cL(q)|_{\vSigma_{\vu\vu}^*} =0$, which leads to:
\begin{align}
{\vSigma_{\vu\vu}^{*-1}} &= \vQ_\Psi -2 \msum_n\nabla_{\vSigma_{\vu\vu}} \EE_{q} [\log p(y_n| f(x_n))]\,.
\end{align}
The term $ \vQ_\Psi $ contributes only to the block-tridiagonal band of the precision.
The posterior prediction for each data point $q(f(x_n))$ only depends on the marginal covariance $\vSigma_{\vv_n\vv_n}$
of the two neighboring inducing states from Eq.~\eqref{eq:marginal-prediction}.
So, each likelihood term for data $y_n$ contributes to the posterior precision at the location $z_\om$, $z_\op$, on which $q(f(x_n))$ depends, which is also in the band.
That is, the optimal $q_{\vu}$ has the same sparsity pattern in the precision as $p_\Psi$. 
So, we choose to parameterise $q_{\vu} = \NN(\vmu_{\vu}, \vQ_{\vu}^{-1})$, by its mean $\vmu_{\vu}$ and by a lower triangular matrix $\vL_{\vu}$ with two block-diagonals which can be interpreted as the Cholesky factor of the block-tridiagonal $\vQ_{\vu}$. This parameterisation results in a $\order{(Md^2)}$ storage footprint instead of $\order{(M^2d^2)}$ had we chosen the more general \textit{mean, covariance} parameterisation.

\subsection{Doubly sparse variational \GP inference: \ssvgp}

We denote by \ssvgp the algorithm that performs sparse variational inference with inducing state-space features while restricting the variational distribution to be in the class of multivariate normal distributions with block-tridiagonal banded precisions. Graphical models summarising the corresponding prior and approximate posterior assumptions are given in Figure~\ref{fig:graph_model}.

\ssvgp has the following computational advantages: 
(1) both the KL divergence and the pairwise marginal posterior predictions of contiguous inducing states (a pairwise Kalman-like smoothing of $q_{\vu}$) can be {evaluated} in $\order(M d^3)$;
(2) Given these pairwise marginals on inducing states, the marginal posterior predictions of function evaluations at the data points $q(f(x_n))$ can be evaluated in parallel in $\order(N d^3)$. 
Overall, the evaluation of the variational loss (and of its gradient) has complexity $\order((N + M)d^3)$ \footnote{Even in this full batch form, this is not always slower than the $\order(Nd^3)$ Kalman smoothing hiding a large constant factor.},
which compares favourably against alternative variational methods as shown in Table~\ref{tab:vgp-complexity}. For efficient implementation of the variational loss and the gradients based on banded precision parameterised Gaussian distributions we refer to \citep{durrande2019banded}.

\ssvgp further inherits two properties of the \svgp approximation. First, marginal posterior predictions $q(f(x_n))$ for each data point can be computed independently which allows to perform stochastic optimisation of an estimator of the loss evaluated on random mini-batches of size $N_b$ of the data, reducing the complexity of an evaluation of the objective to $\order((N_b + M) d^3)$. Second because the inducing inputs $\vz$ are decoupled from the data inputs $\vx$, \ssvgp can be used as a \GP-layer in a deep (compositional) architecture as in \citep{salimbeni2017doubly}. Both properties are not available to alternative state-space approaches to \GP models.

\begin{table*}[tbh]
\small
  \caption{Complexity and capabilities of variational inference algorithms for GP regression.}
  \label{tab:vgp-complexity}
  \centering
  \begin{tabular}{lllcc}
    \toprule
    \rot{Algorithm}     & \rot{Complexity}      & \rot{Storage}  &  \rot{deep $\GP$}  & \rot{minibatch}\\
    \midrule
    VGP \citep{vgp_gpflow} & $\order(N^3)$  & $\order(N^2)$  &  \xmark  &  \xmark \\
    SVIGP \citep{hensman2013gaussian} &  $\order(N_{b} M^2 + M^3)$     & $\order(M^2)$   & \cmark   &  \cmark \\
    VGP (banded) \citep{durrande2019banded}    & $\order(Nd^3)$       & $\order(Nd^2)$  & \xmark  &  \xmark \\
    \ssvgp [this work]    & $\order((N_{b} + M) d^3)$       & $\order(Md^2)$ & \cmark  &  \cmark \\
    \bottomrule
  \end{tabular}
\end{table*}

\subsection{Natural gradient updates}
\label{sec:natgrads}
To learn the variational parameters 
and the parameters of the model we resort to 
gradient based optimisation.
The gradient of the objective w.r.t. a parameter $\vxi$ is defined as $\nabla_{\vxi} \cL = \lim_{\epsilon \to 0} \arg\min_{\boldsymbol\delta} \frac{1}{\epsilon}\cL(\vxi + \boldsymbol\delta) \text{ subject to constraint } ||\boldsymbol\delta|| = \epsilon$. 
Intuitively, it is the direction of steepest descent with respect to the Euclidean norm of $\boldsymbol\delta$, that maximally reduces the loss. 
However, the Euclidean norm can be deceiving when optimising over distributions: small changes in parameters can induce large changes in the distributions, and changing the parameterisation of the distribution typically leads to different optimisation performance. Natural gradients solve this problems by substituting the constraint on the Euclidean norm by 
 $\KL{q_{\vu}(\vxi)}{q_{\vu}(\vxi + \boldsymbol\delta)} = \epsilon$  \citep{amari1998natural}.
Such a constraint can be shown to induce a quadratic norm in the parameter space with curvature given by
the Fisher information matrix $\vF_{\vxi}$. The direction of steepest descent w.r.t. this norm is given by
$\tilde{\nabla}_{\vxi} \cL = (\nabla_{\vxi} \cL) \vF_{\vxi}^{-1}$.

Conveniently, for distributions in the exponential family, the Fisher matrix takes a rather simple form for any parameterisation $\vxi$ \citep{malago2015information}:
\begin{align}
    \vF_{\vxi}=\left(\frac{\dee\boldsymbol\theta}{\dee\vxi}\right)^\top
    \frac{\dee\boldsymbol\eta}{\dee\boldsymbol\theta}\frac{\dee\boldsymbol\theta}{\dee\vxi},
\end{align}
where $\boldsymbol\theta$ are the natural parameters, $\boldsymbol\eta$ the expectation parameters and $\vxi$ the parameterisation of our choice. 
For the variational distribution $q_{\vu}$ defined in Section~\ref{sec:variatonal_inf} these parameters are equivalent to:
\begin{align}
    \boldsymbol\theta &= [\vL_{\vu}\vL_{\vu}^\top\vmu_{\vu},
        -\nicefrac12 \btd[\vL_{\vu}\vL_{\vu}^\top]]\,, \quad \\
    \boldsymbol\eta &= \EE_{q_{\vu}} [\vt(\vu)] = 
        [\vmu_{\vu}, \btd[\vmu_{\vu}\vmu_{\vu}^\top + \vL_{\vu}^{-\top}\vL_{\vu}^{-1}]]\,.
\end{align}
It is clear that the banded structure of the sufficient statistics is reflected in both natural and expectation parameters.
This allows us to derive efficient updates as in \citep{salimbeni18a} using 
the banded operators introduced by \citet{durrande2019banded}:
\begin{align}
    \tilde{\nabla}_{\vxi_{\text[band]}^\top} \cL =
    \frac{\dee\vxi_{\text[band]}}{\dee\boldsymbol\theta_{\text[band]}}
    \frac{\dee\cL}{\dee\boldsymbol\eta_{\text[band]}}.
\end{align}
There is one caveat: the covariance term  $(\vL_{\vu}^{-\top}\vL_{\vu}^{-1})$ in $\boldsymbol\eta$ is a full matrix, but for an efficient update
we need to transfer between $\vxi_{\text[band]} \rightleftarrows \boldsymbol\eta_{\text[band]}$ using only the elements in the band.
This requires a novel operator (the reverse of $\mathbb{I}[\cdot]$ in \citep[Sec. 4.1]{durrande2019banded}), 
which maps the band of a covariance to the Cholesky factor of its banded precision. Detailed transformations between
$\boldsymbol\theta_{\text[band]} \rightleftarrows \vxi_{\text[band]} \rightleftarrows \boldsymbol\eta_{\text[band]}$, along with the algorithm for the operator can be found in the supplementary materials.

In practice, the natural gradient update is recommended when the likelihood is Gaussian since it ensures convergence in very few steps (typically one step, see~\citep{salimbeni18a}). We also observed that it performs well in the non-conjugate case, especially in the first iterations of the optimisation where it can very quickly move to areas of interest. These two properties are illustrated on simple examples in Appendix~\ref{sec:appendix_illustration}.

\section{Experiments}

\subsection{Solar irradiance}
\label{sec:exp:solar}
The aim of this first experiment is to visually illustrate  the predictive power of the proposed methodology on a simple GP regression example. We consider here the solar irradiance
dataset\footnote{https://github.com/jameshensman/VFF.} and choose the model $y(x) = f(x) + \varepsilon$ where $f$ is a centered GP with Mat\'ern \nicefrac{3}{2} covariance and $\varepsilon$ is i.i.d. $\mathcal{N}(0, \sigma^2)$. We compare three (approximate) posteriors for this model: (a) the classic \svgp, (b) the proposed \ssvgp, and (c) the exact GPR posterior. For (a) and (b), a grid of 60 inducing locations is used, and the kernel parameters, the noise variance $\sigma^2$ and the variational parameters are estimated by maximising the ELBO. For (c) the kernel parameters and the noise variance are estimated by maximising the model log-marginal likelihood.

\begin{figure}[ht]
    \centering
        \includegraphics[width=.5\textwidth]{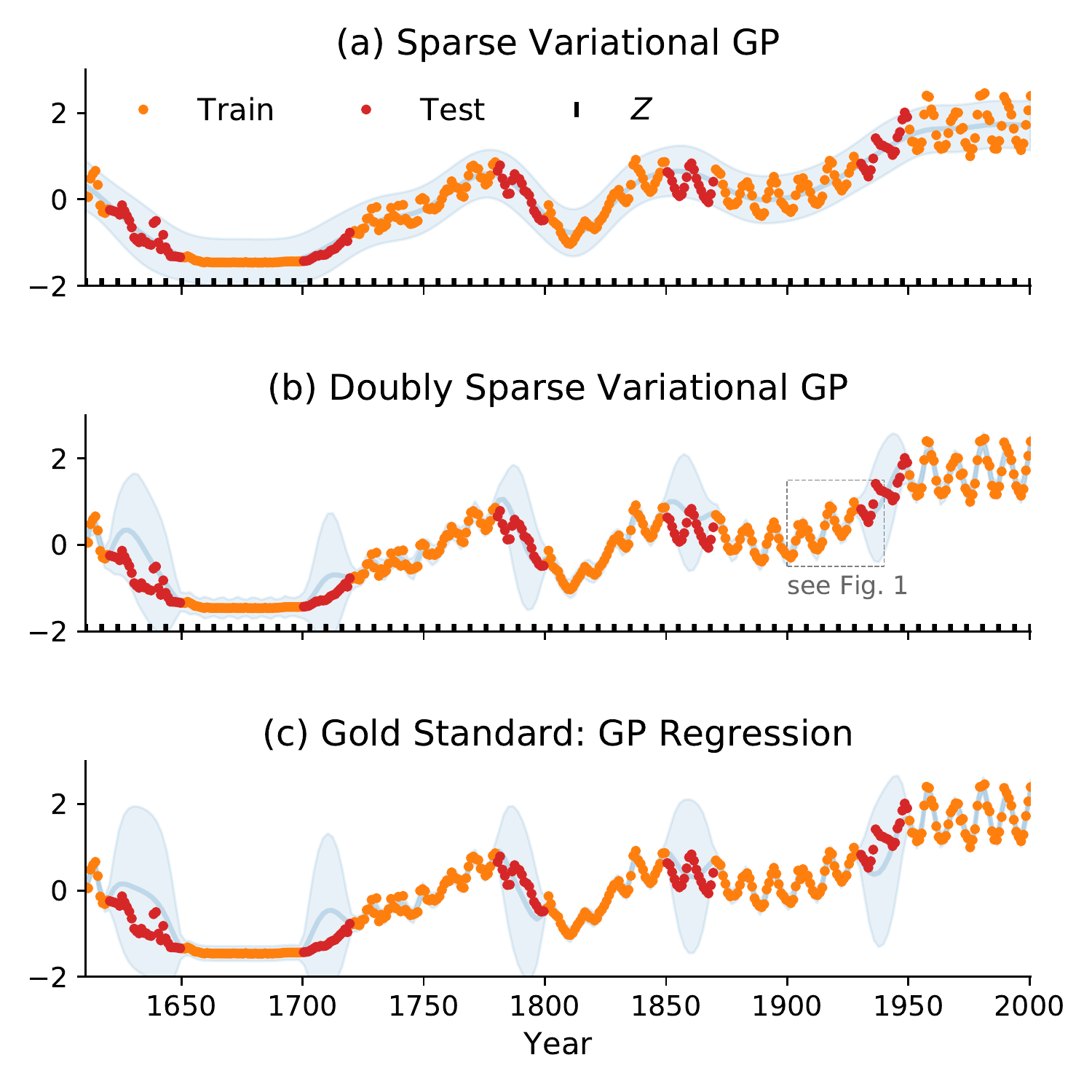}
        \caption{
        Model comparison on the solar dataset. The \svgp (panel a) fails at capturing the high frequency variations, whereas the proposed \ssvgp (panel b) results in predictions that are extremely similar to the ones obtained with exact inference (panel c).
        }
        \label{fig:exp:solar}
\end{figure}

As shown in Figure~\ref{fig:exp:solar}, the proposed method is much more accurate than \svgp on this example, and its predictions are extremely similar to the exact GPR model. One the one hand, this experimental setting may be seen to the advantage of our method since having the same number of inducing inputs means that there are three times more inducing variables in the \ssvgp model. On the other, the computational burden is much smaller for \ssvgp and this added flexibility actually comes with a reduction of the computational cost. This calls for a more thorough investigation that explores how the ELBO of \svgp and \ssvgp compare with respect to the numbers of inducing variables, variational parameters and the execution time. This is what we do in the next section on a larger dataset.

\subsection{Conjugate regression on time series}
\label{sec:exp:conjugate_audio}
In this section we illustrate the computational and storage savings our method entails against the classical \svgp algorithm on a conjugate regression problem (see Appendix \ref{sec:classification_example} for a non-conjugate example). The data consists of an uttered vowel from a female speaker \citep[file \textit{w01ae.wav}]{hillenbrandvoweldata} of length $N=4879$, sampled at $16$kHz. A vowel is a typical quasi-periodic signal where the pitch (or fundamental frequency) is fixed but the repeated pattern varies through time.

\begin{figure*}[ht]
    \centering
        \includegraphics[width=17cm]{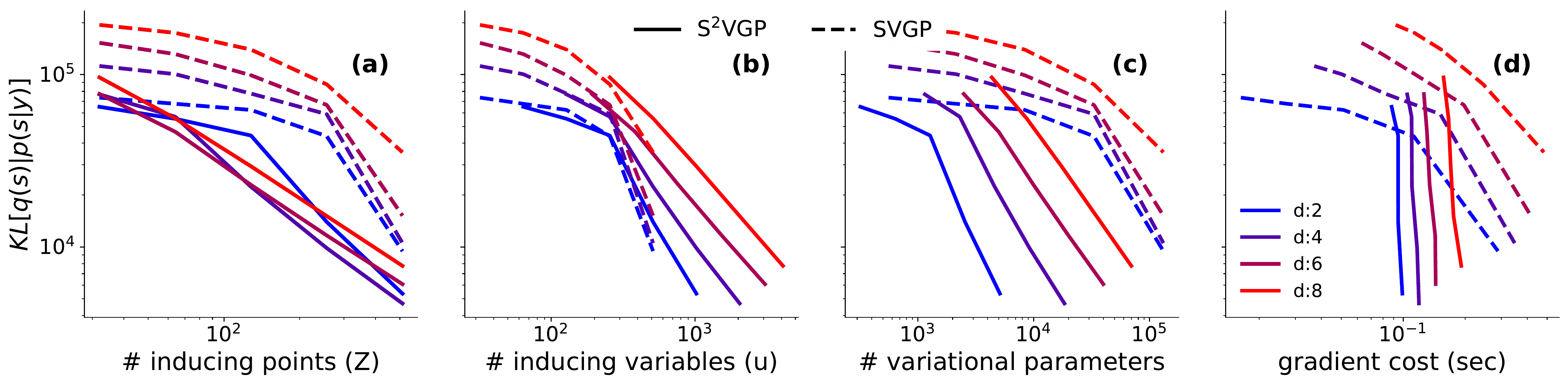}   
        \caption{
        Comparison of sparse variational GP regression using classic inducing points and state-space features on an audio time series (see Fig.~\ref{fig:audio_fit}). Each line corresponds to a given inference method (line style) and model complexity (color) and shows the KL divergence for a an increasing number of inducing points. The proposed method is more accurate for a given number of inducing points, requires less variational parameters to reach a good accuracy, and is extremely fast (especially for large $M$). 
        }
        \label{fig:s2vgp_conjugate}
\end{figure*}

\begin{figure*}[ht]
    \centering
        \includegraphics[width=17cm]{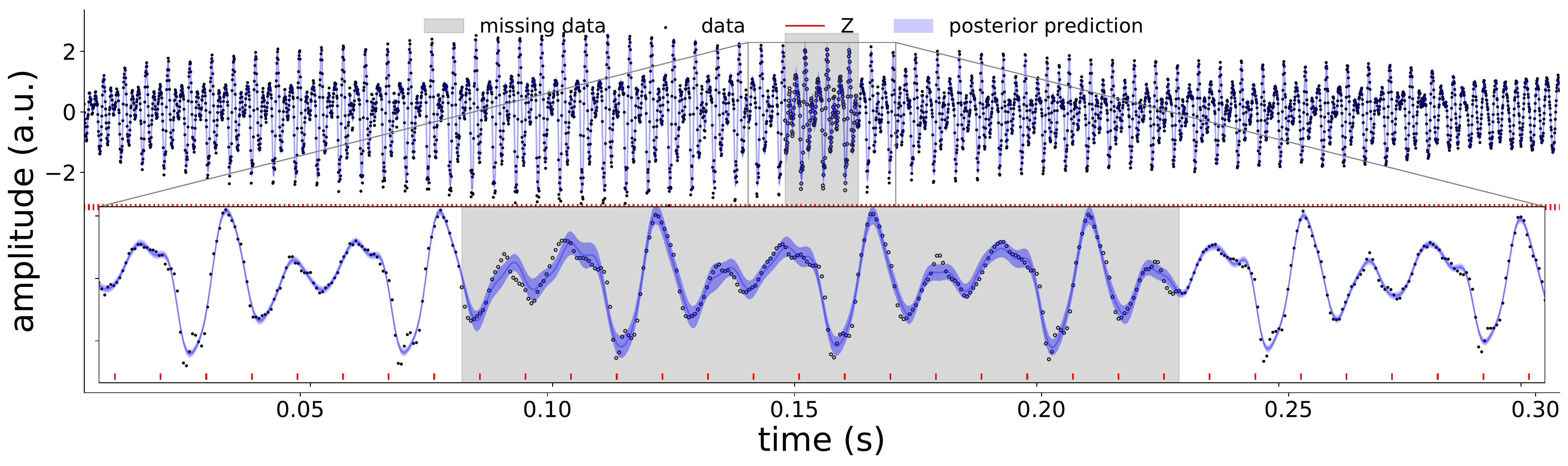}   
        \caption{Vowel waveform with missing data and fit using \ssvgp . Black dots are the data. Blue line and shaded area correspond to the posterior predictions. Red vertical lines are the locations of the inducing points.}
        \label{fig:audio_fit}
    \end{figure*}

We encode our assumptions about this sound by constructing quasi-periodic kernels. Such kernels can be obtained as the product of a periodic kernel $k_p$ and a Mat\'ern \nicefrac{1}{2} kernel $k_{\nicefrac{1}{2}}$ whose lengthscale controls the rate of change of the periodic pattern  \citep{solin2014explicit}.
We construct periodic kernels of varying complexity as weighted sums of cosine kernels in harmonic ratio of frequencies $k^J_p(\tau) = \sum_{j=1}^{J} \gamma_j^2 \cos(2 \pi f_0 j \tau)$ where $f_0$ is the fundamental frequency and $\gamma_j$ controls the magnitude of each harmonic component. Each harmonic increases the state dimension by 2, so the resulting kernel $k^J(\tau) = k^J_p(\tau)k_{1/2}(\tau)$ has a state dimension of $d=2J$.

We then perform approximate inference in settings where we both vary the model complexity (using 1 to 4 harmonics) and the flexibility of the variational distribution by increasing $M$ in powers of 2 (from 16 to 512). For each setting, we optimise the variational parameters and use the divergence $KL[q(\vs(\cdot))|p(\vs(\cdot)|\vy)]$ as a performance metric.
We also compare the execution time of the evaluation of the gradient of $\cL$ with respect to the variational parameters. We use similar implementations of \svgp and \ssvgp as in the previous section. 

Results are displayed in Figure~\ref{fig:s2vgp_conjugate}.  As a function of the number of inducing points $M$, the KL decreases much faster for \ssvgp (a). This is because each inducing state contains more information about the process than an inducing evaluation. Both methods reduce the KL following a similar trend, with a small advantage for \svgp when compared against the actual number of inducing variables (b). However, as summarised in Table~\ref{tab:vgp-complexity}, storage and computational complexities of \ssvgp grow linearly with $M$ but are respectively quadratic and cubic in $M$ for \svgp. Strikingly, the scaling $\bigO((N+M)d^3)$ for the gradient evaluation means the cost of adding extra inducing points in a \ssvgp model is independent of the size of the dataset. For large $N$, it is thus possible to increase the number of inducing variable with very little impact on the computational time since the latter is dominated by the $\bigO(Nd^3)$ term as demonstrated by the vertical lines in (d). Note that this is not the case for \svgp which scales as $\bigO(NM^2 + M^3)$.

A visual illustration of an inference is given in Figure \ref{fig:audio_fit} ($N=4541, M=318$), where we have also removed part of the signal to demonstrate the out-of-sample predictive ability of \ssvgp. Our method correctly interpolates between the periodic patterns at both ends of the missing data region with highest predictive uncertainty in the middle of this region.

\subsection{Additive regression}
\label{sec:exp_additive_regression}
This experiment illustrates three core capabilities of the proposed \ssvgp algorithm: 
(i) it can deal with large datasets; (ii) it allows minibatching with rather large batch-size; (iii) it is not restricted to problems with $1$-dimensional inputs. We also show that the proposed method is competitive in terms of performance.

The airline delay dataset consists of flight details (route distance, airtime, aircraft age, etc.) for every commercial flight in the USA for the year 2008. We use the same $c=8$ covariates $\vx$ as in \citep{hensman2013gaussian} to predict the delay $y$ of the aircraft at landing.

We perform regression from $\RR^c \to \RR$ under the modelling assumption that the delay is additive, i.e.\ $f(\vx) = \msum_{i} f_i(x^{(i)})$, where $x^{(i)}\in \RR$.
We set GP priors over each function $f_i \sim \GP(0, k_i)$, where $k_i$ are Mat\'ern \nicefrac{3}{2} kernels. 
We propose a mean-field approximation to the posterior over processes $q(f_1,\dots,f_c)= \mprod_i q^{(i)}(f_i)$,
where each process is approximated using our \ssvgp parameterisation. Details are given in Appendix~\ref{sec:app:add_reg}. 

We compare the MSE obtained with two different optimisation schemes: the first one optimises all parameters with Adam~\citep{kingma2014adam}, while the second uses natural gradient for the variational parameters and Adam for the remaining ones. All learning rates are set to a constant: $\gamma_{natgrads}=0.01$ and $\gamma_{Adam} = 0.01$. Given the size of the dataset, we use minibatches of 10k points for the two optimisers when $N \geq 10^6$. 

As illustrated in Table~\ref{tab:airline}, natural gradient provides the best performances when the number of observations is small, but as expected it suffers from the minibatching on larger datasets. On the other hand, Adam performs similarly, even when it only has access to sub-samples of the data. The proposed approach has similar accuracy to the state of the art~\citep{hensman2017variational}. A graphical version of the results in this table is given in Appendix (Figure~\ref{fig:airline_mse}).

\begin{table*}[ht!]
\setlength{\tabcolsep}{1.5pt}
  \caption{Predictive mean squared errors (MSEs) and negative log predictive densities
(NLPDs) with one standard deviation on the airline arrival delays experiment.}
  \label{tab:airline}
  \scriptsize
  \centering
  \begin{tabular}{lcccccccc}
    \toprule
    $N$ & \multicolumn{2}{c}{$10k$} & \multicolumn{2}{c}{$100k$} &
    \multicolumn{2}{c}{$1m$}  & 
    \multicolumn{2}{c}{$\approx 6m $}        \\
    \midrule
    & MSE & NLPD
    & MSE & NLPD
    & MSE & NLPD
    & MSE & NLPD \\
    \midrule
    VFF \citep{hensman2017variational} & 
     $0.89 \pm 0.15$ &
     $1.36 \pm 0.09$ &
     $0.82 \pm 0.05$ &
     $1.32 \pm 0.03$ &
     $0.83 \pm 0.01$ &
     $1.336 \pm 0.008$ &
     $0.827 \pm 0.004$ &
     $1.324 \pm 0.003$\\
    \ssvgp (Adam) & 
     $0.96 \pm 0.13$ &
     $1.40 \pm 0.07$ &
     $0.83 \pm 0.07$ &
     $1.33 \pm 0.04$ &
     $0.81 \pm 0.02$ &
     $1.316 \pm 0.010$ &
     $0.822 \pm 0.006$ & 
     $9.22 \pm 20.4$\\
    \ssvgp (Adam+Natgrads) & 
     $0.90 \pm 0.12$ &
     $1.36 \pm 0.07$ &
     $0.81 \pm 0.07$ &
     $1.32 \pm 0.04$ &
     $0.82 \pm 0.02$ &
     $1.321 \pm 0.010$ &
     $0.829 \pm 0.006$ &
     $1.325 \pm 0.004$\\
    \bottomrule
  \end{tabular}
\end{table*}

\subsection{Time warping with deep GPs}
\label{sec:exp_deep}

\begin{figure}[t!]
\centering
        \includegraphics[width=.5\textwidth]{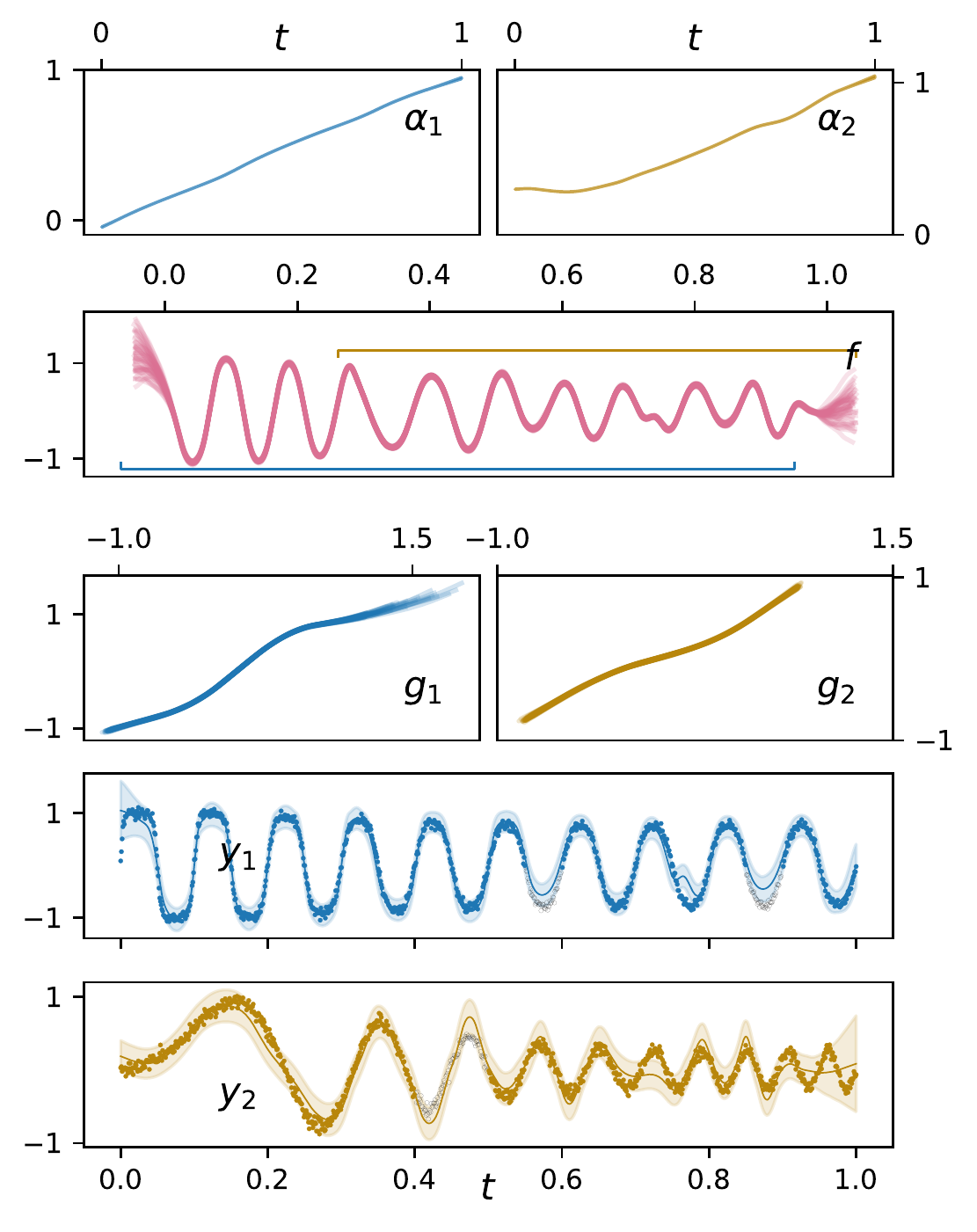}   
 \caption{Data alignment with \ssvgp layers. The top three panels show samples from the inferred functions; the bottom two show the observed data $\vy$ (coloured dots) and posterior predictions for the whole sequence for each output. Black circles indicate missing data.}
        \label{fig:deep_fit}
\end{figure}

In this section we demonstrate the ability of \ssvgp to perform variational inference in a deep-GP model~\citep{damianou2013deep}, where inference is performed by following the approach presented in \cite{salimbeni2017doubly}.
We consider the problem of data alignment and focus on reproducing the results from (\citet[Sec.~4.1]{kaiser2018bayesian}, see Appendix \ref{sec:app:exp_deep}).
\stef{The dataset consists of two times series generated by a three layer model and further corrupted by additive Gaussian noise so that
$\vy_{k} = g_k(f(a_k(t))) + \boldsymbol{\varepsilon}$, where $\boldsymbol{\varepsilon} \sim \NN (0, \sigma^2 \vI) $. The function $f$ is a sine wave shared across the two observed series indexed by $k=\{1, 2\}$.
The functions $a_k$ are time-warping functions, with $a_1$ being the identity and $a_2$ a quadratic function; while $g_k$ are output distortions applied to $f$, with $g_1$ the hyperbolic tangent and $g_2$ the identity.
Parts of the observed sequences have been removed at different locations to assess the generalisation performance of the model.}
\vince{Compared to the setting of \cite{kaiser2018bayesian}, we double both the frequency of the true $f$ and the number of observations.}

The goal here is to infer all five functions $a_1, a_2, f, g_1, g_2$ under the true model structure.
We place GP priors on all five functions with Mat\'ern \nicefrac{3}{2} kernels. We further introduce linear mean functions on all priors to
avoid pathologies while propagating samples through the layers \citep{salimbeni2017doubly}.
\stef{We use $M = [50, 100, 50]$ inducing points at each layer with the inducing inputs placed on a linear grid and kept fixed (all functions within a layer share the inducing input locations).
We use natural gradients to learn the variational parameters of the approximate distributions in each layer, with the learning rate set to $\gamma_{natgrads} = 0.001$.}
The hyper-parameters of the kernel, the mean functions and the likelihood noise are learnt using Adam~\citep{kingma2014adam} with exponential decaying learning rate (initialised from $\gamma_{Adam} = 0.001$).

Results of the inference are shown in Figure~\ref{fig:deep_fit}. \stef{The inferred $a_k$ and $g_k$ functions (first and third row) share the same characteristics as the corresponding ground truth functions. The function $f$ is also recovered with the correct dampening (second row).
More interestingly, we see that
due to the time warping of the first layer, the model has successfully learnt to reconstruct each observed sequence from different sections of $f$.
Finally, in the lower two panels we report the model predictions highlighting how the learnt model in rightfully more uncertain in the missing data region.} Uncertainty decomposition across layers is discussed in Appendix~\ref{sec:app:exp_deep}.

\section{Discussion and conclusions}

The proposed doubly sparse variational Gaussian processes combines the variational sparse approximations for Gaussian processes with the state-space representation of the process.
It inherits its appealing tractability from the variational approach, but has the representative power and computational scalability of state-space representations. Unlike other state-space \GP methods, it is readily applicable to deep \GP settings and supports mini-batch stochastic training. 
We showed that our framework can be used to approximate functions with more than one input variable while preserving the computational gain of state-space models.

To ease the optimisation of our variational objective, we derived natural gradient updates for the class of multivariate normal distribution with banded precisions. Although the objective is non-convex, this leads to few shot inference in the conjugate setting and empirically improves optimisation in non-conjugate settings.
To further improve the applicability of \ssvgp, different sub-optimal variational parameterisations could be used for $q_{\vu}$ leading to better behaved objectives or additional scalability improvement, albeit at the cost of reduced expressivity. Another route of improvement could consist in making a further steady-state approximation to the posterior to reduce the scaling with the \GP state dimension from cubic to quadratic as in \citep{solin2018infinite}. 
As in \citep{nickisch2018state}, further computational gains could be achieved by using interpolations when we compute the SSM parameters.



\bibliographystyle{plainnat}
\bibliography{main}

\newpage
\onecolumn

\input{appendix.tex}

\bibliographystyleApp{plainnat}
\bibliographyApp{main}

\end{document}

%% file: graphs/graphical_model_small.tex
\begin{center}
\begin{tikzpicture}[thick,scale=0.8, every node/.style={scale=0.9}]

\node (x1) [dummy] {$x_1$};
\node (z1) [dummy, right = .5\ndi of x1, red] {$z_1$};
\node (x2) [dummy, right = 2.1\ndi of x1] {$x_2$};
\node (x3) [dummy, right = 3.6\ndi of x1] {$x_3$};
\node (z2) [dummy, right = 5.\ndi of x1, red] {$z_2$};
\node (x4) [dummy,  right = 6.9\ndi of x1] {$x_4$};
\node (z3) [dummy,  right = 8.4\ndi of x1, red] {$z_M$};
\node (xN) [dummy,  right = 10\ndi of x1] {$x_N$};

\newcommand \yf {1.2\ndi};
\newcommand \yu {2.9\ndi};
\newcommand \yline {.35\ndi};
\newcommand \yprior {4.5\ndi};

\path let \p0 = (x1) in node[func] (f1) at (\x0,\yf ) {$\vec s_1$};
\path let \p0 = (x2) in node[func] (f2) at (\x0,\yf ) {$\vec s_2$};
\path let \p0 = (x3) in node[func] (f3) at (\x0,\yf ) {$\vec s_3$};
\path let \p0 = (x4) in node[func] (f4) at (\x0,\yf ) {$\vec s_4$};
\path let \p0 = (xN) in node[func] (fN) at (\x0,\yf ) {$\vec s_N$};

\path let \p0 = (z1) in node[state] (s1) at (\x0,\yu) {$\vec u_1$};
\path let \p0 = (z2) in node[state] (s2) at (\x0,\yu) {$\vec u_2$};
\path let \p0 = (z3) in node[state] (s3) at (\x0,\yu) {$\vec u_M$};

\draw (-\ndia,\yline) -- (14.5\ndi,\yline);

\draw[thick] let \p{f}=(f1) in (\x{f},\yline-.1\ndia)--(\x{f},\yline+.1\ndia);
\draw[thick] let \p{f}=(f2) in (\x{f},\yline-.1\ndia)--(\x{f},\yline+.1\ndia);
\draw[thick] let \p{f}=(f3) in (\x{f},\yline-.1\ndia)--(\x{f},\yline+.1\ndia);
\draw[thick] let \p{f}=(f4) in (\x{f},\yline-.1\ndia)--(\x{f},\yline+.1\ndia);
\draw[thick] let \p{f}=(fN) in (\x{f},\yline-.1\ndia)--(\x{f},\yline+.1\ndia);

\draw[thick,red] let \p{f}=(s1) in (\x{f},\yline-.1\ndia)--(\x{f},\yline+.1\ndia);
\draw[thick,red] let \p{f}=(s2) in (\x{f},\yline-.1\ndia)--(\x{f},\yline+.1\ndia);
\draw[thick,red] let \p{f}=(s3) in (\x{f},\yline-.1\ndia)--(\x{f},\yline+.1\ndia);

\newcommand \opac {0.2};

\newcommand \dso {.5};




\tikzdots (f4) -- (fN); 
%
%

\draw [graphmod sf] (s1) -- (f2) node[left,pos=0.5]{} ;
\draw [graphmod sf] (s2) -- (f2) node[left,pos=0.5]{} ;

\draw [graphmod sf, opacity=\opac] (s1) -- (f1) node[left,pos=0.5]{} ;
\draw [graphmod sf, opacity=\opac] (s1) -- (f3) node[left,pos=0.5]{} ;
\draw [graphmod sf, opacity=\opac] (s2) -- (f3) node[left,pos=0.5]{} ;
\draw [graphmod sf, opacity=\opac] (s2) -- (f4) node[left,pos=0.5]{} ;
\draw [graphmod sf, opacity=\opac] (s3) -- (f4) node[left,pos=0.5]{} ;
\draw [graphmod sf, opacity=\opac] (s3) -- (fN) node[left,pos=0.5]{} ;

\draw [graphmod ss] (s1) -- (s2) node[left,pos=0.5]{} ;
\draw [graphmod ss] (s2) -- (s3) node[left,pos=0.5]{} ;

\path let \p0 = (x1) in node[dummy] (a) at (\x0-1.5\ndi, \yprior ) {$(a)$};
\path let \p0 = (x1) in node[dummy] (b) at (\x0-1.5\ndi, .5*\yu+.5*\yf){$(b)$};

\path let \p0 = (x1) in node[func] (s1p) at (\x0,\yprior ) {$\vec s_1$};
\path let \p0 = (x2) in node[func] (s2p) at (\x0,\yprior ) {$\vec s_2$};
\path let \p0 = (x3) in node[func] (s3p) at (\x0,\yprior ) {$\vec s_3$};
\path let \p0 = (x4) in node[func] (s4p) at (\x0,\yprior ) {$\vec s_4$};
\path let \p0 = (xN) in node[func] (sNp) at (\x0,\yprior ) {$\vec s_N$};
\path let \p0 = (z1) in node[state] (u1p) at (\x0,\yprior) {$\vec u_1$};
\path let \p0 = (z2) in node[state] (u2p) at (\x0,\yprior) {$\vec u_2$};
\path let \p0 = (z3) in node[state] (u3p) at (\x0,\yprior) {$\vec u_M$};

\draw [graphmod sf, black] (s1p) -- (u1p) node[left,pos=0.5]{} ;
\draw [graphmod sf, black] (u1p) -- (s2p) node[left,pos=0.5]{} ;
\draw [graphmod sf, black] (s2p) -- (s3p) node[left,pos=0.5]{} ;
\draw [graphmod sf, black] (s3p) -- (u2p) node[left,pos=0.5]{} ;
\draw [graphmod sf, black] (u2p) -- (s4p) node[left,pos=0.5]{} ;
\draw [graphmod sf, black] (s4p) -- (u3p) node[left,pos=0.5]{} ;
\draw [graphmod sf, black] (u3p) -- (sNp) node[left,pos=0.5]{} ;


\end{tikzpicture} 
\end{center}

%% file: appendix.tex
\begin{appendices}
\setlength{\abovedisplayskip}{5pt}
\setlength{\belowdisplayskip}{5pt}

\section{Details on the doubly sparse variational Gaussian process}

\subsection{Conditional for Markovian Gaussian processes}
\label{sec:appendix_conditional}

We consider a stationary Markovian GP with state dimension $d$ and denote by $(\vu_{-}, \vs, \vu_{+})$ its evaluation on the triplet $(z_\om, t, z_\op)$.
We here detail the derivation of $p(\vs| \vv=[\vu_{-}, \vu_{+}])$

\subsubsection*{Derivation from the joint precision}
\begin{align*}
p(\vs| \vu_{-},  \vu_{+}) &\propto p(\vs| \vu_{-})p(\vu_{+}|\vs) \\
&\propto \NN (\vs; \vA_{\om,t} \vu_{-}, \vQ_{\om,t}) \NN(\vu_{+};\vA_{t,\op} \vs, \vQ_{t,\op})\\
&\propto  \exp -\frac{1}{2}
\left[  ||\vs - \vA_{\om,t} \vu_{-}||^2_{\vQ_{\om,t}^{-1}}  + ||\vu_{+} - \vA_{t,\op} \vs||^2_{\vQ_{t,\op}^{-1}} \right]\\
&\propto  \exp -\frac{1}{2}\big[ 
     \vs^\top \underset{\vT^{-1}}{\underbrace{(\vQ_{\om,t}^{-1} + (\vA_{t,\op})^\top \vQ_{t,\op}^{-1} \vA_{t,\op})}} \vs  - 2 \vs^\top 
     \underset{\vM = [\vM_1, \vM_2]}{\underbrace{\begin{bmatrix} \vQ_{\om,t}^{-1} \vA_{\om,t}, &  \vA^\top_{t,\op} \vQ_{t,\op}^{-1}  \end{bmatrix}}} \vv  \big] \\
&\propto  \exp -\frac{1}{2}\left[ \vs^\top \vT^{-1} \vs - 2 \vs^\top  \vM \vv   \right]  = \NN(\vs; \vP \vv, \vT) 
\end{align*}

with
\begin{align*}
\vT &=  (\vQ_{\om,t}^{-1} + \vA^\top_{t,\op} \vQ_{t,\op}^{-1} \vA_{t,\op})^{-1} \; \text{(Woodbury identity)}\\
&=  \vQ_{\om, t} -  \vQ_{\om,t} \vA^\top_{t,\op} ( \vQ_{t,\op} + \vA_{t,\op} \vQ_{\om,t}\vA_{t,\op}^\top)^{-1}\vA_{t,\op}  \vQ_{\om,t}\\
&=  \vQ_{\om, t} -  \vQ_{\om,t} \vA^\top_{t,\op} \vQ_{\om,\op}^{-1}  \vA_{t,\op}  \vQ_{\om,t}\\
\end{align*}
and
$    \vP = [\vP_1, \vP_2] = \vT \vM = [ \vT\vM_1,  \vT\vM_2] $ given by\\
\begin{align*}
\vP_1 &= (\vQ_{\om, t} -  \vQ_{\om,t} \vA^\top_{t,\op} \vQ_{\om,\op}^{-1}  \vA_{t,\op}  \vQ_{\om,t}) \vQ_{\om,t}^{-1} \vA_{\om,t}  \\
&= \vA_{\om,t}  -  \vQ_{\om,t} \vA^\top_{t,\op} \vQ_{\om,\op}^{-1}   \vA_{\om,\op}  \\
\vP_2 &= (\vQ_{\om, t} -  \vQ_{\om,t} \vA^\top_{t,\op} \vQ_{\om,\op}^{-1}  \vA_{t,\op}  \vQ_{\om,t}) \vA^\top_{t,\op} \vQ_{t,\op}^{-1}  \\
&=\vQ_{\om, t} \vA^\top_{t,\op} \vQ_{t,\op}^{-1} -   \vQ_{\om, t} \vA^\top_{t,\op}\vQ_{\om,\op}^{-1} (\vQ_{\om,\op} - \vQ_{t,\op})  \vQ_{t,\op}^{-1} \; \text{(Woodbury identity)}\\
&=  \vQ_{\om, t}\vA^\top_{t,\op}\vQ_{\om,\op}^{-1}
\end{align*}

\subsubsection*{Derivation from the joint covariance}

Another derivation using the covariance approach.
One can write down the joint density 
\[p(\vs, \vv) =
\NN\left(
\begin{bmatrix} \vmu_{\vs}\\ \vmu_{\vv} \end{bmatrix},
\begin{bmatrix} \vSigma_{\vs\vs} & \vSigma_{\vs\vv}\\
\vSigma_{\vv\vs} & \vSigma_{\vv\vv}    \end{bmatrix}\right)\]
with 
\begin{align*}
\vmu_{\vs} &= \vmu_{\vv} = \boldsymbol{0}\\
\vSigma_{\vs\vs} &= \vP_0\\
\vSigma_{\vv\vv} &= \begin{bmatrix} \vP_0 & \vA_{\om,\op}^T\vP_0\\
\vP_0 \vA_{\om,\op} & \vP_0    \end{bmatrix}\\
\vSigma_{\vs\vv} &= \begin{bmatrix} \vA_{\om,t}^T\vP_0 , & \vA_{\op,t}^T\vP_0 \end{bmatrix}
\end{align*}

and get

\[
p(\vs| \vv) =
\NN\left(
 \vmu_{\vs}+ \vSigma_{\vs\vv}\vSigma_{\vv\vv}^{-1} (\vv  -  \vmu_{\vv})\\,
 \vSigma_{\vs\vs} - \vSigma_{\vs\vv}\vSigma_{\vv\vv}^{-1} \vSigma_{\vv\vs}   \right)
 \]

Both implementations reveal the overall $\bigO(d^3)$ scaling of the conditional statistics.

\subsection{Sampling from the variational posterior process}
\label{sec:appendix_sample}

We here describe a method to jointly sample from the posterior $q(\vs(\cdot))$ at inputs $\vx$.
Such a sample can be obtained by first sampling from the prior process at inputs $[\vx, \vz]$:  
\begin{align}
\vs_p, \vu_p \sim p(\vs(\vx),\vs(\vz))
\end{align}
Then sampling from the marginal posterior at $\vz$: 
\begin{align}
\vu_q \sim q(\vs(\vz))
\end{align}
And finally construct:
\begin{align}
\vs = \vs_{p} + E[\vs(\vx)|\vs(\vz)=\vu_q - \vu_p],
\end{align}
which is a sample from the marginal posterior $q(\vs(\vx))$.

This methods allows to generate samples in complexity $\bigO((N+M)d^3)$, which is the time required to jointly sample $s_p, u_p$. It was used to produce the posterior samples in the deep-\GP experiment.

\section{Multivariate Gaussian distributions with banded precision: parameterisations, link functions and natural gradients}

Here we present different parameterisations of a multivariate Gaussian distribution with block- tridiagonal precision matrices along with the link functions between them and describe how we use these to compute natural gradient updates of our variational objective. 

\subsection{Distribution parameterisations and link functions}

In Section~\ref{sec:variatonal_inf} we have defined the variational distribution approximating the posterior on inducing states to be
\begin{align}
    q_{\vu} = \NN(\vmu_{\vu}, \vQ_{\vu}^{-1}), \quad \vQ = \vL_{\vu}\vL_{\vu}^\top,
\end{align}
where $\vQ_{\vu}$ denotes the precision matrix with block-tridiagonal structure and $\vL_{\vu}$ the Cholesky factor of the precision.
We denote with $\vxi: \{ \vm_{\vu}, \vL_{\vu}\}$ the above parameterisation. 
In the following table we present the identities that allow us to transfer back and forth from
the default parameterisation $\vxi$ to the natural parameters 
$\boldsymbol\theta:\{\boldsymbol\theta_1, \boldsymbol\theta_2\}$
and to the expectation parameters
$\boldsymbol\eta:\{\boldsymbol\eta_1, \boldsymbol\eta_2\}$.

\begin{table}[tbh]
\label{tab:gauss_param}
\centering
  \caption{Transformations between the different parameterisations of the Gaussian distribution with block-tridiagonal precision.}
  \centering
  \begin{tabular}{ccll}
    \toprule
    Transformation &
    \multicolumn{1}{c}{Original parameterisation}
    & \multicolumn{2}{l}{\qquad\quad Resulting parameterisaton}\\
    \midrule
 $\vxi \to \boldsymbol\theta $   
 & $\vmu_{\vu}$,
\quad $\btd[\vL_{\vu}]$ 
 &  $\boldsymbol\theta_1 = \vL_{\vu}\vL_{\vu}^\top\vmu_{\vu}$,
 & $\boldsymbol\theta_2 = -\nicefrac{1}{2}\btd[\vL_{\vu}\vL_{\vu}^\top]$ \\ 
 $\boldsymbol\theta \to  \vxi$  
 & $\boldsymbol\theta_1$,
 \quad $\btd[\boldsymbol\theta_2]$
 & $\vmu_{\vu} = (-2 \boldsymbol\theta_2)^{-1}\boldsymbol\theta_1$,
& $\vL_{\vu}= \btd[\chol[-2 \boldsymbol\theta_2]]$ \\ 
 $\vxi \to \boldsymbol\eta$
 & $\vmu_{\vu}$,
 \quad$\btd[\vL_{\vu}]$ 
 &$\boldsymbol\eta_1 = \vmu_{\vu}$,
 &$\boldsymbol\eta_2 = \btd[\vL_{\vu}^{-\top}\vL_{\vu}^{-1} + \vmu_{\vu}\vmu_{\vu}^\top$]\\ 
 $\boldsymbol\eta\to  \vxi$ 
 & $\boldsymbol\eta_1$,
 \quad$\btd[\boldsymbol\eta_2]$
 & $\vmu_{\vu} = \boldsymbol\eta_1$, 
 &$\vL_{\vu} = \btd[\chol[ (\boldsymbol\eta_2 - \boldsymbol\eta_1 \boldsymbol\eta_1^\top)^{-1}]]$ \\ 
    \bottomrule
  \end{tabular}
\end{table}


Note that the expectation parameter $\boldsymbol\eta_2$ is a full matrix since it
involves the inverse of a banded matrix, which is
not necessarily banded.
However, since the sufficient statistics of the
distribution are $\vt(\vu) = [\vu, \btd[\vu \vu^\top]]$,
we only need to compute the elements in the band.

\subsection{Natural gradient update}

When maximising our objective $\cL(\vxi)$ with respect to $\vxi$, the parameters of our variational distribution, we perform a sequence of natural gradient updates:
\[
\vxi_{t+1} = \vxi_{t} - \gamma_t \tilde{\nabla}_{\vxi}\cL|_{\vxi=\vxi_t}, \quad
\tilde{\nabla}_{\vxi}\cL|_{\vxi=\vxi_t} = \vF_{\vxi}^{-1}\nabla_{\vxi^\top} \cL|_{\vxi=\vxi_t}
\]

In an exponential family with \emph{natural parameters} $\boldsymbol\theta$ and \emph{expectation parameters} $\boldsymbol\eta$, the Fisher information is given by
\[
\vF_{\vxi}=\left(\frac{\dee\boldsymbol\theta}{\dee\vxi}\right)^\top\frac{\dee\boldsymbol\eta}{\dee\boldsymbol\theta}\frac{\dee\boldsymbol\theta}{\dee\vxi}
\]
As shown in \citeApp{salimbeni18a}, this leads to 
\[
\tilde{\nabla}_{\vxi^\top}\cL = \frac{\partial \vxi}{\partial \boldsymbol\theta}\frac{\partial \cL}{\partial \boldsymbol\eta^\top},
\]
which is a Jacobian-vector product allowing for an efficient implementation using automatic differentiation libraries. The computation of  $\frac{\partial \cL}{\partial \boldsymbol\eta}$ is achieved using the chain rule: 
$\frac{\partial \cL}{\partial \boldsymbol\eta} = \frac{\partial \cL}{\partial \vxi}\frac{\partial \vxi}{\partial \boldsymbol\eta}$.

%
%
%
%
%
%
%
%
%
%
%
%
%
%
%

\subsection{Inverse of the subset inverse, and its reverse mode derivatives}

A banded positive semi-definite (PSD) matrix $Q$ has a Cholesky factor $L_Q$  that is lower triangular with the same lower bandwidth. However, its inverse $Q^{-1}$ is in most cases dense.
If one is only interested in computing the entries of $Q^{-1}$ that are located in the band of $Q$ (denoted by band$_Q[\cdot]$), efficient \emph{subset inverse} algorithms are available \citepApp{durrande2019banded}.
We are now interested in the mathematical inverse of this operation, which we call \emph{reverse} to avoid confusion with the matrix inverse operation.
\[
L_Q 
\xleftrightarrows[\text{subset inverse}]{\text{reverse subset inverse}} \text{band}_{Q}[Q^{-1}]
 \]

We consider symmetric matrices with lower bandwidth $r$
\[
Q = \left( 
\begin{array}{cccccc}
q_{11} & \dots & q_{1r} & & \dots& 0\\
\vdots & \ddots & \ddots & \ddots  & & \vdots \\
q_{r1} & \ddots &\ddots & \ddots&\ddots &  \\
& \ddots & \ddots & \ddots  &  \ddots & q_{n-r,n} \\
\vdots &&\ddots&\ddots&\ddots& \vdots\\
0 & \dots & &q_{n,n-r} & \dots& q_{n,n}  
\end{array}  
\right) .
\]

Such a matrix has banded Cholesky factor
\[
L_Q = \left( 
\begin{array}{cccccc}
l_{11} &  &  & & \dots& 0\\
\vdots & \ddots &  &   & & \vdots \\
l_{r1} & \ddots &\ddots & & &  \\
& \ddots & \ddots & \ddots  &   & \\
\vdots &&\ddots&\ddots&\ddots& \\
0 & \dots & &l_{n,n-r} & \dots& l_{n,n}  
\end{array}  
\right) .
\]


Computing the \emph{subset inverse} is $\order{(n r^2)}$ \citepApp{takahashi1973formation}. We present below an algorithm that performs the \emph{reverse subset inverse} operation with complexity $\order{(n r^2)}$.

\subsubsection{Derivation of the forward evaluation $\mathrm{band}_Q[C] \to L_Q$}

The following derivation shows how to compute, for each index $i$, the column $L_{i:i+r,i}$ given the sub-block $C_{i:i+r,i:i+r}$ independently.

To simplify notations, we introduce the following intervals $\nn=[1:i-1]$, $\oo = [i:i+r]$  and $\pp = [i+r+1:n]$. We denote by $C$ the full covariance $C=Q^{-1}= L^{-T}L^{-1}$.

First, we have that the sub-covariance $C_{\oo\pp,\oo\pp}$ only depends on $L_{\oo\pp,\oo\pp}$
\begin{align}
L^{-1} = 
\begin{bmatrix}
    L_{\nn,\nn}       & 0  \\
    L_{\oo\pp,\nn}       & L_{\oo\pp,\oo\pp}
\end{bmatrix}^{-1}
= \begin{bmatrix}
    L_{\nn,\nn}^{-1}       & 0  \\
    -L_{\nn,\nn}^{-1}L_{\nn,\oo\pp}L_{\oo\pp,\oo\pp}^{-1}       & L_{\oo\pp,\oo\pp}^{-1}
\end{bmatrix}
\implies C_{\oo\pp, \oo\pp} =  L_{\oo\pp,\oo\pp}^{-T}L_{\oo\pp,\oo\pp}^{-1} 
\end{align}
Second, we have that the following expression for sub-covariance $C_{\oo,\oo}$
\begin{align}
 [L_{\oap,\oap}]^{-1} &=
\begin{bmatrix}
    L_{\oo,\oo}       & 0  \\
    L_{\pp,\oo}       & L_{\pp,\pp}
\end{bmatrix}^{-1}
= \begin{bmatrix}
    L_{\oo,\oo}^{-1}       & 0  \\
    -L_{\pp,\pp}^{-1}L_{\pp,\oo}L_{\oo,\oo}^{-1}       & L_{\pp,\pp}^{-1}
\end{bmatrix}
\end{align}
\begin{align}
\implies C_{\oo,\oo} = [L_{\oap,\oap}^{-T}L_{\oap,\oap}^{-1}]_{1:r,1:r} &=  L_{\oo,\oo}^{-T} L_{\oo,\oo}^{-1} +  L_{\oo,\oo}^{-T} L_{\pp,\oo}^T  L_{\pp,\pp}^{-T}  L_{\pp,\pp}^{-1}  L_{\pp,\oo}  L_{\oo,\oo}^{-1} 
\end{align}
Using, the matrix inversion lemma, we get the following expression for $C_{\oo,\oo}^{-1}$
\begin{align}
C_{\oo, \oo}^{-1}& = [L_{\oap,\oap}^{-T}L_{\oap,\oap}^{-1}]_{1:r,1:r}^{-1} \\
&= L_{\oo,\oo}L_{\oo,\oo}^T - L_{\oo,\oo}L_{\pp,\oo}^T(L_{\pp,\pp}L_{\pp,\pp}^T + L_{\pp,\oo}L_{\pp,\oo}^T)^{-1}L_{\pp,\oo}L_{\oo,\oo}^T \label{eq:inverse_cov}
\end{align}
By construction, the first column of $L_{\pp\oo}$ is out of the matrix band (it is a null vector). Because $L_{\oo,\oo}$ is lower triangular, the product $L_{\pp\oo}L_{\oo,\oo}^T$ also has a null vector as its first column, and so has the last term in Eq.~\ref{eq:inverse_cov}.

Therefore, keeping only the first column, we end up with the identity
\begin{align}
[C_{\oo, \oo}^{-1}]_{:,1} =  L_{\oo,\oo}[L_{\oo,\oo}^T]_{:,1} = L_{\oo,i}L_{i,i},
\end{align}
which is a system of $r$ equations with $r$ unknown $L_{\oo,i}$, that we can solve analytically getting first $L_{i,i}= \sqrt{[C_{\oo, \oo}^{-1}]_{1,1}}$ then $ L_{\oo,i} = [C_{\oo, \oo}^{-1}]_{:,1}/L_{i,i}$

This derivation is summarised in Algorithm \ref{alg:inv_inv_general}:
\begin{algorithm}[h]
\caption{Reverse subset inverse for banded matrices}\label{alg:inv_inv_general}
\begin{algorithmic}[1]
\Procedure{Rev\_Subset\_Inv}{$C$}\Comment{ $C = \text{band}_{Q}[Q] \in \RR^{n\times n}$, $Q=LL^T$ of bandwidth $r$}
\State $L \gets 0$
\For{$i \in [0, \dots, n-1]$}
\State $c^{(i)} \gets C_{i:i+r, i:i+r}$ \Comment{extract symmetric sub-block $(r \times r)$ at $i$}
\State $v^{(i)} \gets (c^{(i)})^{-1}e, \quad  e = [1,0,\dots,0]\in \RR^r$ \Comment{select first column of $(c^{(i)})^{-1}$}
\State $l^{(i)} \gets v^{(i)}/ \sqrt{v^{(i)}_{1}}$ 
\State $L_{i:i+r,i} \gets l^{(i)}$ 
\EndFor
\State \textbf{return} $L$\Comment{Cholesky factor of $Q$}
\EndProcedure
\end{algorithmic}
\end{algorithm}

\subsubsection{Derivation of the reverse mode differentiation}

We manually derive the reverse mode differentiation of the \emph{reverse inverse subset} algorithm introduced in the previous section.

We refer the reader to \citeApp{giles2008collected} for a brief introduction to reverse mode differentiation and to \citeApp{durrande2019banded} for derivations of reverse mode differentiation of the \emph{subset inverse} algorithm for banded matrices.

In a nutshell,  in a chain $A \to B \to ... \to c \in \RR$ with $A, B$ matrices, the reverse mode derivative of operation of $f:A \to B$ is the operation propagating the reverse mode \emph{sensitivity} $ \bar{B} = \frac{dc}{dB}$ into sensitivity $\frac{dc}{dA}$. Given the differential identity $dc = \sum_{ij} \frac{\partial c}{\partial B_{ij}}dB_{ij} = Tr[\bar{B}^T dB]$ and the general differential relation at $A$: $dB = X_A  dA Y_A$, it follows that  $dc = Tr[Y_A ^T X_A  dA ]$, therefore we identify $\bar{A}^T = Y_A \bar{B}^T X_A$.

Algorithm \ref{alg:inv_inv_general} being parallel, we can compute the contribution of each column of $L$ to $C$ separately. 
%
First we relate $l^{(i)} $ to $v^{(i)}$:

\begin{align*}
dl^{(i)} &=  \underbrace{\frac{1}{\sqrt{v^{(i)}_{1}}}\begin{bmatrix} 
    1 - l^{(i)}_1 \big/ \Big(2 \sqrt{v^{(i)}_{1}}\Big)   & & &  \\
     - l^{(i)}_2 \big/ \Big(2 \sqrt{v^{(i)}_{1}}\Big)    & 1 &&  \\
    \vdots &  &\ddots &\\
    - l^{(i)}_n \big/ \Big(2 \sqrt{v^{(i)}_{1}}\Big)   &      &  & 1 
    \end{bmatrix}}_{H_i} dv^{(i)} = H_i  dv^{(i)}\, .
\end{align*}
We identify $\bar{v}^{(i)}$:
\begin{align*}
df = \sum_i tr(i{\bar{l}^{(i)}}{}^T dl^{(i)}) = \sum_i tr({\bar{l}^{(i)}}{}^T H_i  dv^{(i)}) \implies {\bar{v}^{(i)}}{}^T= {\bar{l}^{(i)}}{}^T H_i
\end{align*}
Then we relate $v^{(i)}$ to $c^{(i)}$, 
\begin{align*}
v^{(i)} =  \big(c^{(i)}\big)^{-1} e, \implies dv^{(i)} = - \big(c^{(i)}\big)^{-1} dc^{(i)}  \big(c^{(i)}\big)^{-1} e\\
\end{align*}
And we identify $\bar{c}^{(i)}$:
\begin{align*}
df &= \sum_i tr(\bar{v}^{(i)}{}^T dv^{(i)}) = \sum_i tr(- \big(c^{(i)}\big)^{-1} e \bar{v}^{(i)}{}^T   \big(c^{(i)}\big)^{-1} dc^{(i)}  )\\
&\implies  \bar{c}^{(i)}{}^T= - \big(c^{(i)}\big)^{-1} e^{(i)}\bar{v}^{(i)}{}^T   \big(c^{(i)}\big)^{-1}\\
\end{align*}
Putting everything together, we have
\begin{align*}
 \bar{c}^{(i)}{}^T &= - \big(c^{(i)}\big)^{-1} e \bar{l}^{(i)T} H_i   \big(c^{(i)}\big)^{-1}\\
\end{align*}

Algorithm \ref{alg:rev_mode_inv_inv_general} summarises the derivations of the reverse mode sensitivity $\bar{C}$ of the \emph{reverse subset inverse} operation.

\begin{algorithm}[H]
\caption{Reverse mode sensitivity: reverse of subset inverse}\label{alg:rev_mode_inv_inv_general}
\begin{algorithmic}[1]
\Procedure{Grad\_Rev\_Subset\_Inv}{$\bar{L}, C$}\Comment{$\bar{L}=  \frac{df}{dL}$ is $n\times n$}
\State $\bar{C} \gets 0$
\For{$i \in [0, \dots, n-1]$}
\State $c^{(i)} \gets C_{i:i+r, i:i+r}$ \Comment{extract symmetric sub-block $(r \times r)$ at $i$}
\State $\bar{l}^{(i)} = \bar{L}_{i:i+r,i}$
\State $\bar{c}^{(i)} = - \big(c^{(i)}\big)^{-1} e \bar l^{(i)}{}^T H_i   \big(c^{(i)}\big)^{-1}$
\State $\bar{C}_{i:i+r,i:i+r} \gets \bar{C}_{i:i+r,i:i+r} +  \bar{c}^{(i)}$ \Comment{add to sensitivity $\bar{C}$}
\EndFor
\State \textbf{return} $\bar{C}$ \Comment{ $\bar{C} = \frac{df}{dC}$}
\EndProcedure
\end{algorithmic}
\end{algorithm}

\subsection{Fast implementation}

In Algorithms~\ref{alg:inv_inv_general}~and~\ref{alg:rev_mode_inv_inv_general}, we computed the $(c^{(i)})^{-1}e$ independently for each $i$. This can be achieved by first computing the Cholesky factors $s^{(i)}$ of each $c^{(i)}$ and then solving $v^{(i)} = (s^{(i)})^{-T}(s^{(i)})^{-1}e$. 
The direct Cholesky factorization of all $c^{(i)}$ would incur a total complexity of $\order{(nr^3)}$. However one can use a recursive algorithm achieving the same goal in complexity $\order{(nr^2})$.

Given the Cholesky factor $s^{(i)}$ of $c^{(i)}$, we can compute the Cholesky factor of $c^{(i-1)}$ as follows:
\begin{align}
s^{(i-1)} 
&= \chol(C_{i-1:i+r,i-1:i+r})_{1:r,1:r} \\
&= \chol \left(\begin{bmatrix}
    C_{i-1,i-1}       & C_{i-1:i+r,i-1}^T  \\
    C_{i-1:i+r,i-1}       & s^{(i)}s^{(i)T}
\end{bmatrix}\right)_{1:r,1:r}
\end{align}
with
\begin{align}
&\chol \left(\begin{bmatrix}
    C_{i-1,i-1}       & C_{i-1:i+r,i-1}^T  \\
    C_{i-1:i+r,i-1}       & s^{(i)}s^{(i)T}
\end{bmatrix}\right)\\
=& \begin{bmatrix}
    \sqrt{C_{i-1,i-1}}       & 0  \\
    C_{i-1:i+r,i-1}/\sqrt{C_{i-1,i-1}}     & \chol(s^{(i)}s^{(i)T} - C_{i-1:i+r,i-1}C_{i-1:i+r,i-1}^T/C_{i-1,i-1} )
\end{bmatrix} \label{eq:recursive_chol}
\end{align}

The Cholesky factor $s^{(i-1)}$ is then readily obtained by removing the last row and column of the matrix in Eq.~\ref{eq:recursive_chol}.
The bottom right entry of the matrix in Eq.~\ref{eq:recursive_chol} corresponds to a Cholesky \emph{downdate} that can be computed with cost $r^2$ \citepApp{seeger2004low}, so that, starting from index $n$, all Cholesky factors $s^{(1)},\dots, s^{(n)}$ can be computed with complexity $\order{(nr^2)}$.

\section{Experimental details}
\label{sec:appendix_exp_details}

\subsection{Illustration of the efficiency of the natural gradient update}
\label{sec:appendix_illustration}
\begin{figure}[h!]
    \centering
        \includegraphics[width=14cm]{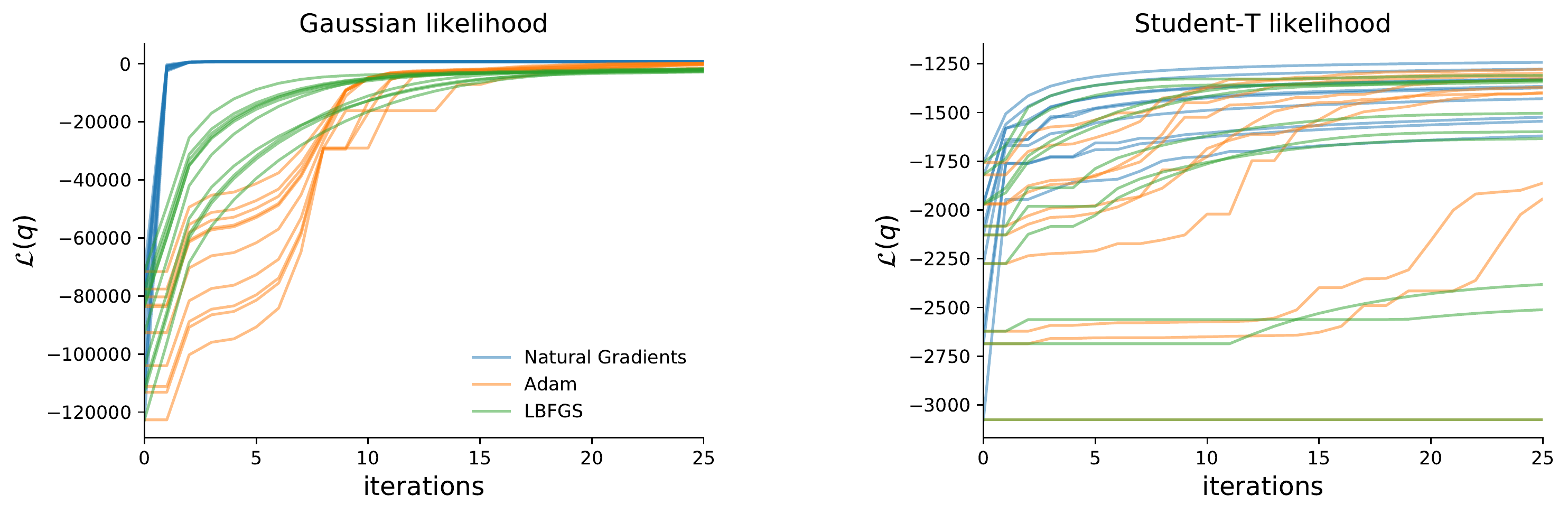}   
        \caption{Optimisation traces for the ELBO of our \ssvgp method. Three optimisers (Natural gradients, Adam and LBFGS) are compared on 10 datasets that are generated at random. The left pannel correspond to a dataset and a model with a conjugate likelihood, whereas the right one is for a non conjugate likelihood.}
        \label{fig:natgrad_vs_adam}
    \end{figure}

We describe in this section a simple experiment to compare the behaviour of various optimisers. Given a regular grid X of $10^3$ points equally spaced on $[0, 1]$ and a centred GP $f$ with a Mat\'ern 3/2 covariance (unit variance and length-scale $\ell = 0.1$), we generate two datasets at random as follow:
\begin{align}
    \vy &= f(X) + \varepsilon \text{\qquad with } \varepsilon_i \sim \mathcal{N}(0, 0.01)\\ 
    \vz &= f(X) + \tau \text{\qquad with } \tau_i \sim \mathcal{T}(df=1)\, . 
\end{align}
Note that the first model has a conjugate likelihood whereas the second one does not. We can then fit \ssvgp models with 50 inducing points (fixed to a regular grid on $[0, 1]$). We show in Figure~\ref{fig:natgrad_vs_adam} the optimisations traces we obtained when optimising the variational parameters (all other model parameters being fixed to their nominal values), for different samples of the datasets. It can be seen that natural gradients provide a striking advantage in the conjugate case but that it also behaves favourably, especially in the first few iterations, in the non-conjugate case.

\subsection{Details on Section~\ref{sec:exp:conjugate_audio}: conjugate regression on time-series}

\begin{figure}[ht]
    \centering
        \includegraphics[width=17cm]{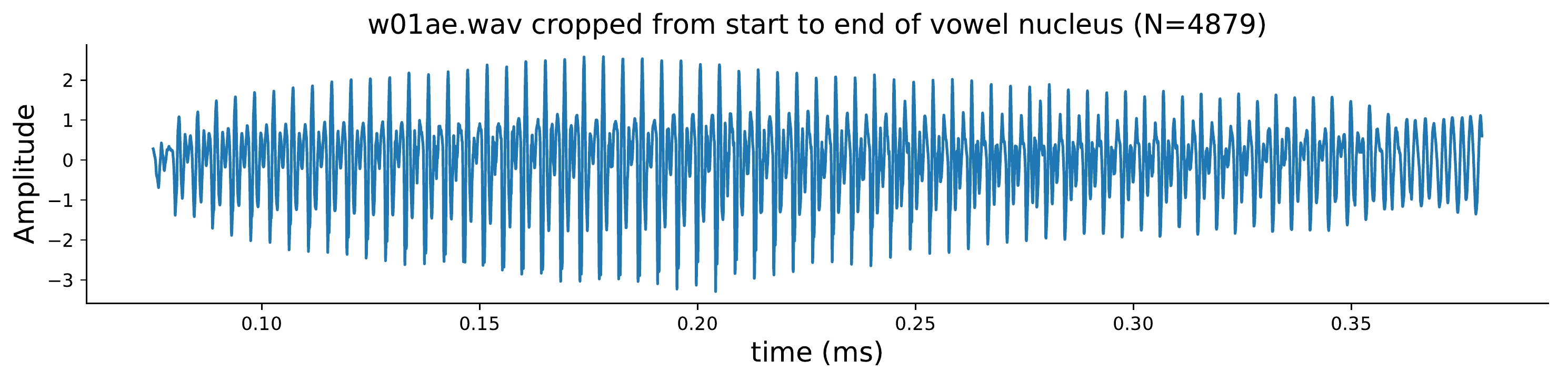}   
        \caption{Audio time-series used in Section~\ref{sec:exp:conjugate_audio}}
        \label{fig:audio_file}
    \end{figure}

The full sound waveform used in this experiment is shown in Figure~\ref{fig:audio_file}(top).

To model this signal, we used the following stationary kernels that have an equivalent SDE representations of state-dimension $2J$, where $J$ is the number of harmonic components:
\[
k^J(\tau) = k_{Mat\nicefrac{1}{2}}(\tau) \left( \msum_{j=1}^{J} \gamma_j^2 \cos(2\pi f_0 j \tau)  \right)
\]
where $f_0$ denotes the fundamental frequency of the pitched sound. The variance of the Mat\'ern\nicefrac{1}{2} kernel was set to one and its only free parameter is its length-scale $\ell$.
We used a Gaussian likelihood with variance $\sigma^2$.
Since we focused on inference, all parameters were initially fitted to the data by maximising the marginal likelihood available in closed form in this conjugate setting.

We increased the number of inducing points in powers of $2$, placing them on an homogenous grid from the start to the end of the time support. Inducing points locations were not learned.

For both \svgp and \ssvgp, we only learned the variational parameters.
We used $LBFGS$ for both \svgp and \ssvgp.

\subsection{Details on Section~\ref{sec:exp_additive_regression}: additive regression}
\label{sec:app:add_reg}

The generative model is as follow:
\begin{align}
f_i &\sim \GP(0, k_i),\quad k_i = \text{Matern}_{\nicefrac{3}{2}}(\sigma^2_i, \ell_i),\\
y_k &= \sum_{i=1}^c f_i(x^{(i)}_k) + \epsilon_k,\quad \epsilon_k \sim \NN(0, \sigma^2),\quad x^{(i)}\in \RR.
\end{align}
We propose a mean-field approximation to the posterior over processes $g(f_1,\dots,f_c)= \mprod_i q^{(i)}(f_i)$ as in \citepApp{adam2016scalable}.
where each process is approximated using our doubly sparse parameterisation with inducing states
$\vu^{(i)}=f_i(\vz^{(i)})$ evaluated at component specific inputs $\vz^{(i)}$.

\begin{figure*}[ht]
    \centering
        \includegraphics[width=10cm]{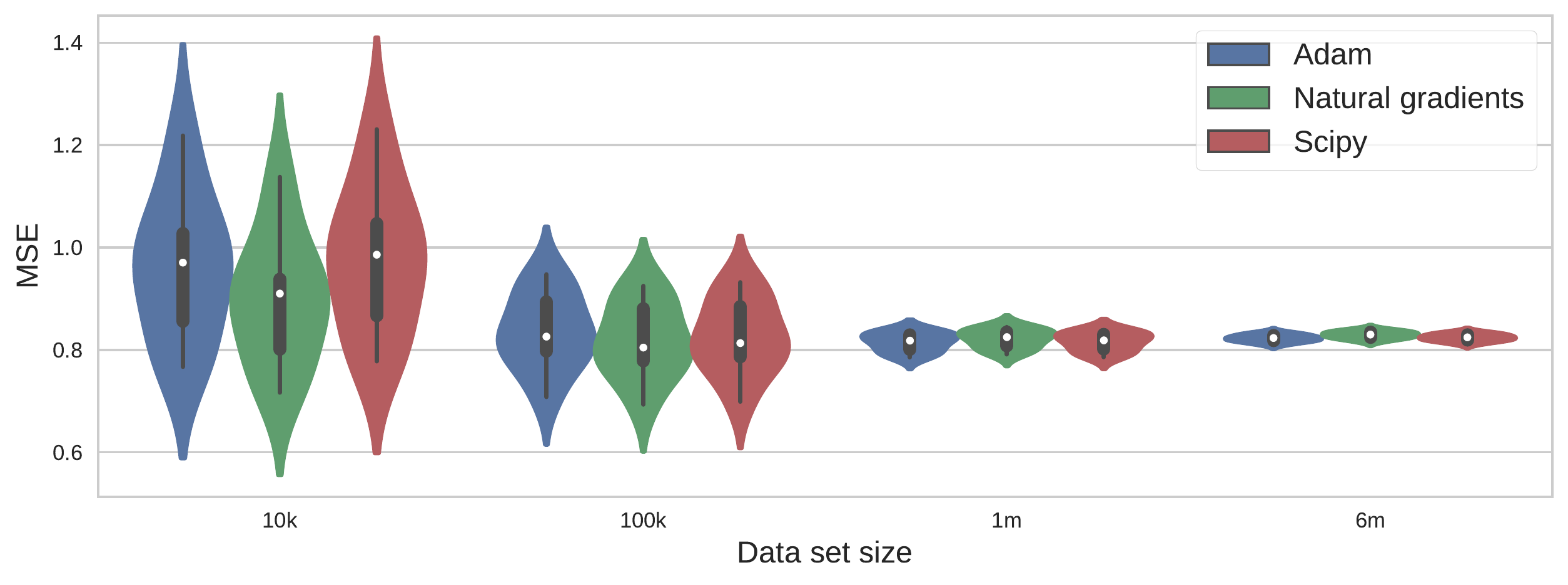}   
        \caption{
         Comparison of predictive MSE on the airline delays dataset when training \ssvgp with various optimisers (the distribution of errors is across the 10 splits).
        }
        \label{fig:airline_mse}
\end{figure*}

\subsection{Details on Section~\ref{sec:exp_deep}: time warping with deep Gaussian
process}
\label{sec:app:exp_deep}

\subsubsection*{Setting}
The function $f$ is a sine wave of the form $f(x) = (0.75 (1 - \tanh(10 * 2\pi x / 15)) + 0.25)  \sin(10 * 2\pi  x)$ and is shared across the two observed series. The functions $a_k$ are time-warping functions, with $a_1(x) = x$ and $a_2(x) = x^2$. The functions $g_k$ are output distortions with $g_1(x) = \tanh(x)$ and $g_2(x) = x$. To generate the two time series we uniformly sample 1000 points in $[0, 1]$ and subsequently pass them to the 2-layer model that we described. The Gaussian additive noise has standard deviation $\sigma = 0.05$. We removed observations in the intervals $[0.55, 0.6]$ \& $[0.85, 0.9]$ for the first time series and in the interval of $[0.40, 0.50]$ for the second time series.

\subsubsection*{Uncertainty decomposition across layers in the deep GP experiment}

\begin{figure}[ht]
    \centering
\begin{tabular}{cc}
 \includegraphics[width=80mm]{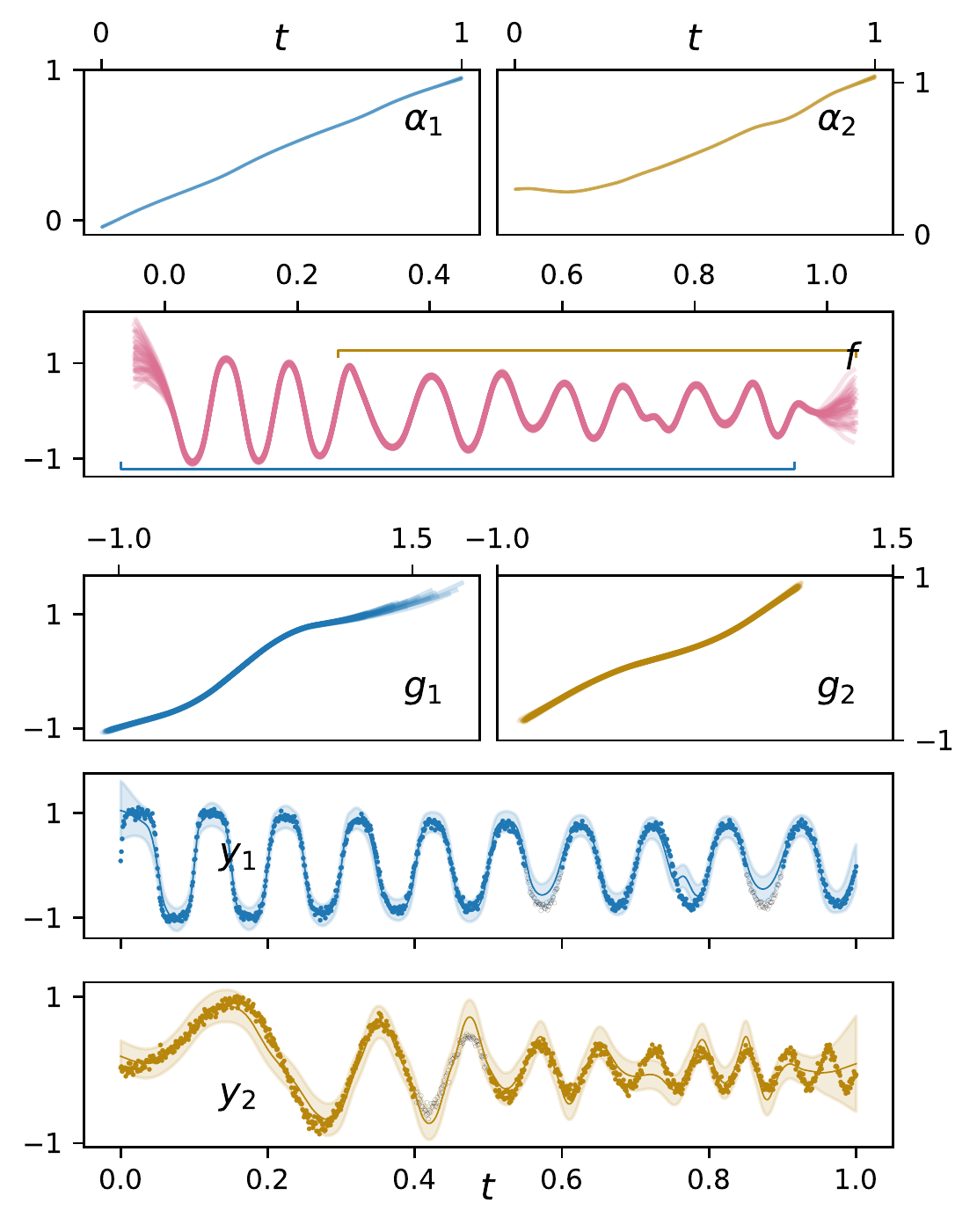} & 
 \includegraphics[width=80mm]{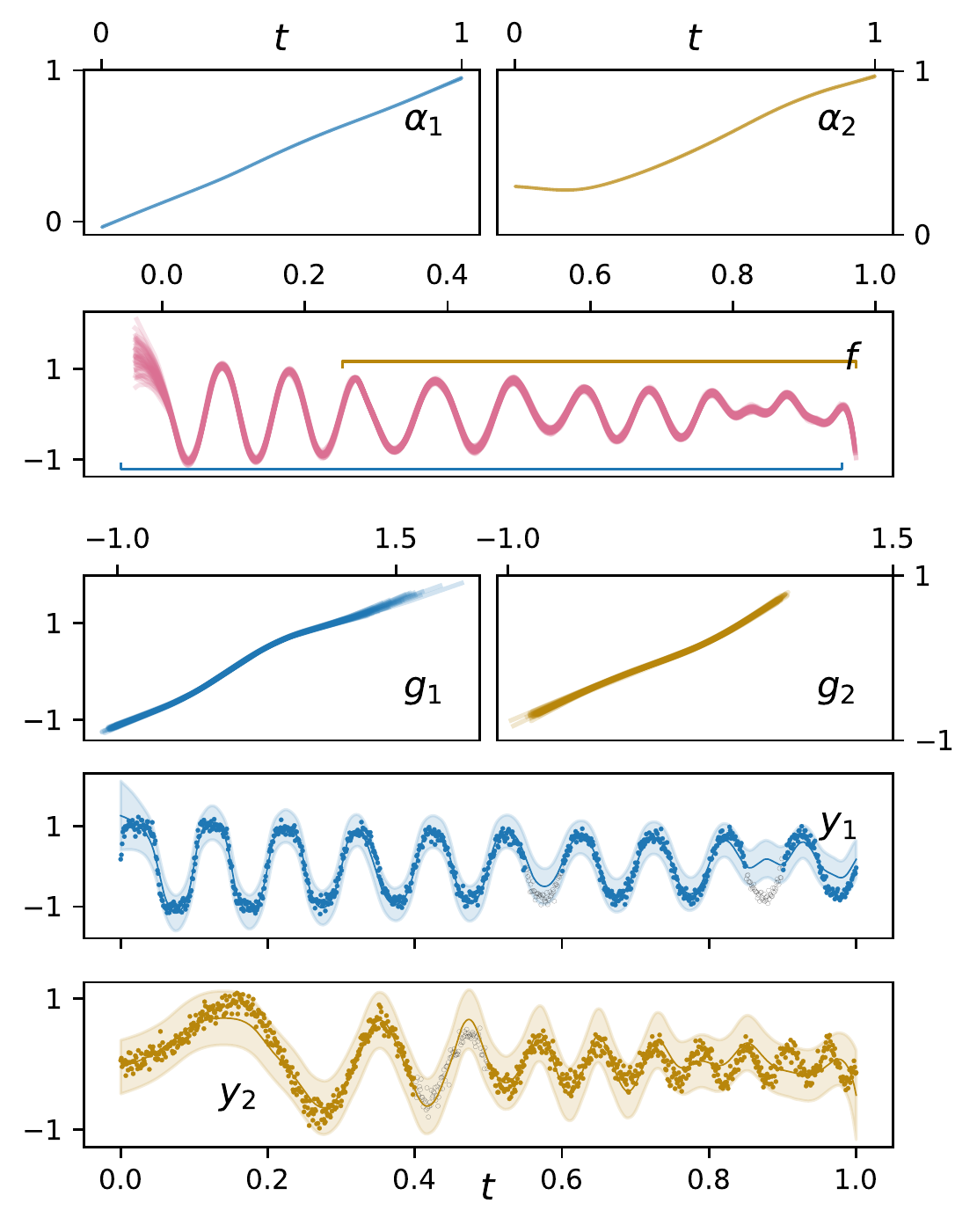} \\
\end{tabular}
\caption{
         Data alignment with \ssvgp layers. \textbf{Left}: original experiment from Section \ref{sec:exp_deep} with $\sigma=0.05$. \textbf{Right}: same experiment with  $\sigma=0.1$.}
\label{fig:deep_high_noise}
\end{figure}

As can be seen in Figure \ref{fig:deep_fit}, there is almost no uncertainty in the first layer. The reason for this is two fold. First, we initialised the kernels on the first layer to have very long lengthscales (initial value of 4), and small variance (initial value of 0.01) to bias the inference towards smooth functions. Second, the low posterior uncertainty in the intermediate layers is a known consequence of variational approach used in~{\citetApp{salimbeni2017doubly},} where the variational distribution factorises across layers (we use the same approximation). This pathology has been recently explored in \citetApp{ustyuzhaninov2019compositional} and can be remedied by explicitly imposing a conditional dependency between the layers in the variational distribution.
We conducted the same experiment with a higher observation noise level ($\sigma=0.1$ instead of $\sigma=0.05$) and report the result in Figure \ref{fig:deep_high_noise}. Changing the noise level has no effect to the uncertainty in the intermediate layers but has a significant effect in the model’s ability to learn the correct functions, as the model prefers to explain these attributes by measurement noise.

%

\section{Empirical comparison to alternative SSM based approximate inference methods}
\label{sec:classification_example}

We empirically compare our \ssvgp approach to alternative methods to perform approximate inference in GP models based on their state space representation. 

\begin{figure}[ht]
\centering
\includegraphics[width=160mm]{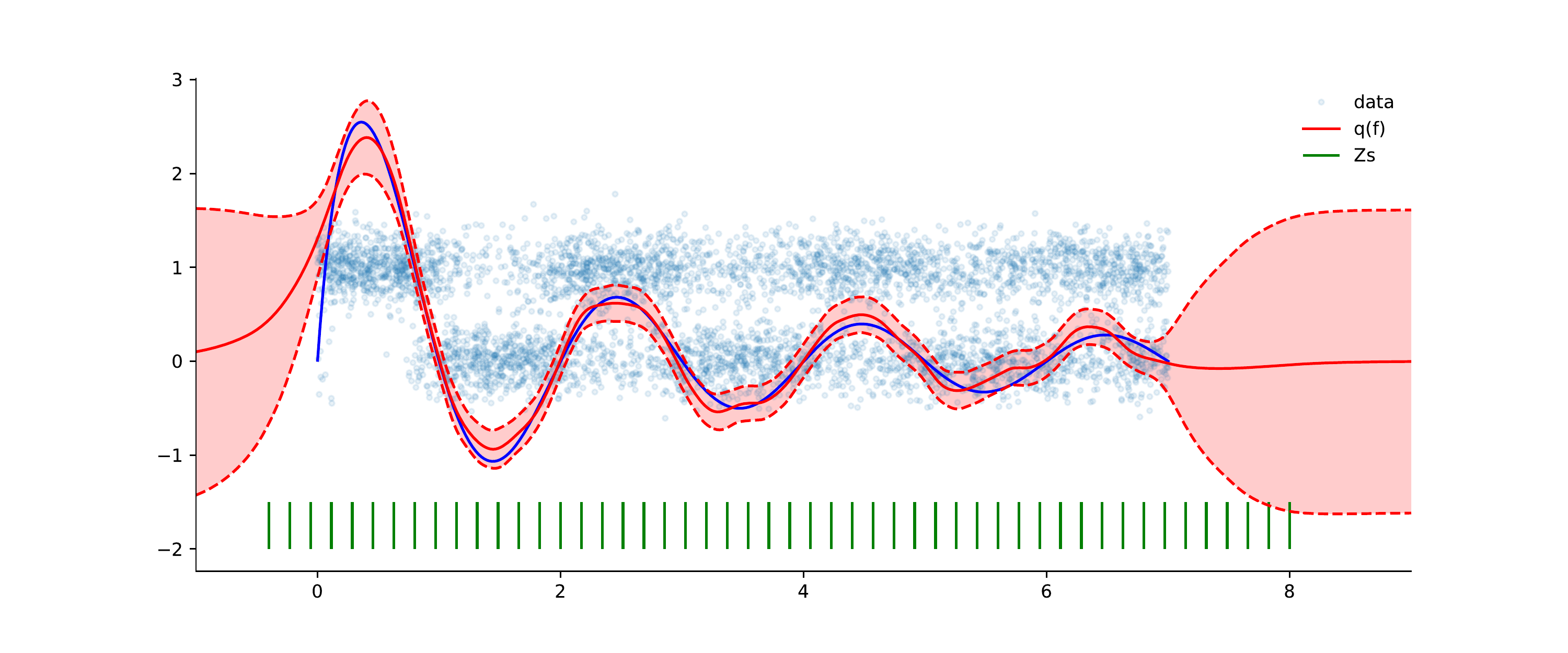} 
        \caption{
        Classification data (N=5000) and \ssvgp fit. The ground truth used to generate the data is shown in blue. Blue dots represent the binary data (with additional noise introduced for visibility). The posterior process is shown in red. Inducing point locations are shown in green.
        }
        \label{fig:classification_example}
\end{figure}

We consider a simple classification task similar from the GPMLv4.2 toolbox demo \texttt{gpml-matlab-master/doc/demoState.m} with  N=5000 (see  Figure \ref{fig:classification_example}) and using Matern\nicefrac{3}{2} kernel with fixed hyperparameters. For the \ssvgp method, we choose $M=50$ inducing points on a homogenous grid and we use gradient based optimisation (L-BFGS). We run inference and report the NLPDs and execution time:

\begin{center}
\begin{tabular}{ lcc } 
    \toprule
Algorithm &           NLPD &               Time (s)     \\
    \midrule
\ssvgp[M=50] (ours)  &  0.586 $\pm$ 0.013    &    3.38 $\pm$ 0.98 \\
Laplace (gpml)      &  0.586 $\pm$ 0.013    &    7.55 $\pm$ 0.15 \\
ADF/EP (gpml)       & 0.586 $\pm$ 0.013     &   10.585 $\pm$ 0.014 \\
VB (gpml)           & 0.587 $\pm$ 0.013     &  39.530 $\pm$ 0.064 \\
    \toprule
\end{tabular}
\end{center}

For \ssvgp, we run gradient based optimisation (using L-BFGS). We are equally accurate as the other methods yet much faster.

\end{appendices}